\documentclass[journal]{IEEEtran}
\usepackage{mathrsfs}
\usepackage{multirow}
\usepackage{booktabs}
\usepackage{hyperref}
\usepackage{color}

\usepackage{colortbl}
\definecolor{indian1}{rgb}{0,0,.6125}
\definecolor{indian2}{rgb}{0,0.0769,1}
\definecolor{indian3}{rgb}{0,0.6154,1}
\definecolor{indian4}{rgb}{0,0.0769,1}
\definecolor{indian5}{rgb}{0.5385,1,0.9231}
\definecolor{indian6}{rgb}{1,0.9230,0}
\definecolor{indian7}{rgb}{1,0.4615,0}
\definecolor{indian8}{rgb}{0.923,0,0}
\definecolor{indian9}{rgb}{0,0.2308,0}
\definecolor{indian10}{rgb}{0,	0.692307692307692,	0}
\definecolor{indian11}{rgb}{0.153846153846154,	0,	0.846153846153846}
\definecolor{indian12}{rgb}{0.615384615384615,	0,	0.384615384615385}
\definecolor{indian13}{rgb}{0,	0,	1}
\definecolor{indian14}{rgb}{0.538461538461538,	0,	1}
\definecolor{indian15}{rgb}{1,	0,	1}
\definecolor{indian16}{rgb}{1,	0,	0.538461538461538}

\definecolor{paviaU1}{rgb}{0,	0,	0.571428571428571}
\definecolor{paviaU2}{rgb}{0,	0.428571428571429,	1}
\definecolor{paviaU3}{rgb}{0.428571428571429,	1,	0.571428571428571}
\definecolor{paviaU4}{rgb}{1,	0.714285714285714,	0}
\definecolor{paviaU5}{rgb}{0.857142857142857,	0,	0}
\definecolor{paviaU6}{rgb}{0,	0.714285714285714,	0}
\definecolor{paviaU7}{rgb}{0.428571428571429,	0,	0.571428571428571}
\definecolor{paviaU8}{rgb}{0.142857142857143,	0,	1}
\definecolor{paviaU9}{rgb}{1,	0,	1}

\ifCLASSINFOpdf
  \usepackage[pdftex]{graphicx}
  \graphicspath{{../pdf/}{../jpeg/}}
\else
  \usepackage[dvips]{graphicx}
  \graphicspath{{../eps/}}
\fi

\usepackage{algorithm} 
\usepackage{algorithmic} 
\usepackage{amsmath,amsthm,amssymb}

\begin{document}

%
\title{SuperPCA: A Superpixelwise PCA Approach for Unsupervised Feature Extraction of Hyperspectral Imagery}

\author{Junjun~Jiang,~\IEEEmembership{Member,~IEEE,~}
        Jiayi~Ma,~\IEEEmembership{Member,~IEEE,~}
        Chen Chen,~\IEEEmembership{Member,~IEEE,~}
        Zhongyuan Wang,~\IEEEmembership{Member,~IEEE,~}
        Zhihua, Cai,
        and Lizhe~Wang,~\IEEEmembership{Senior Member,~IEEE}
\thanks{The research was supported by the National Natural Science Foundation of China (61501413, 61503288, 61671332, 61773355).}
\IEEEcompsocitemizethanks{


\IEEEcompsocthanksitem J. Jiang is with the School of Computer Science and Technology, Harbin Institute of Technology, Harbin 140001, China. E-mail: junjun0595@163.com.
\IEEEcompsocthanksitem J. Ma is with the Electronic Information School, Wuhan University, Wuhan 430072, China. E-mail: jyma2010@gmail.com.
\IEEEcompsocthanksitem C. Chen is with the Center for Research in Computer Vision, University of Central Florida (UCF), Orlando, FL 32816-2365 USA. E-Mail: chenchen870713@gmail.com. 
\IEEEcompsocthanksitem Z. Wang is with the School of Computer, Wuhan University, Wuhan 430072, China. E-mail: wzy\_hope@163.com.
\IEEEcompsocthanksitem Z. Cai and L. Wang are with the School of Computer Science, China University of Geosciences, Wuhan 430074, China. E-mail: zhcai@cug.edu.cn; lizhe.wang@gmail.com.
}

}

\markboth{ }%
{Shell \MakeLowercase{\textit{\emph{et al.}}}: Bare Demo of IEEEtran.cls for Journals}

\maketitle

\begin{abstract}
As an unsupervised dimensionality reduction method, principal component analysis (PCA) has been widely considered as an efficient and effective preprocessing step for hyperspectral image (HSI) processing and analysis tasks. It takes each band as a whole and globally extracts the most representative bands. However, different homogeneous regions correspond to different objects, whose spectral features are diverse. It is obviously inappropriate to carry out dimensionality reduction through a unified projection for an entire HSI. In this paper, a simple but very effective superpixelwise PCA approach, called SuperPCA, is proposed to learn the intrinsic low-dimensional features of HSIs. In contrast to classical PCA models, SuperPCA has four main properties. (1) Unlike the traditional PCA method based on a whole image, SuperPCA takes into account the diversity in different homogeneous regions, that is, different regions should have different projections. (2) Most of the conventional feature extraction models cannot directly use the spatial information of HSIs, while SuperPCA is able to incorporate the spatial context information into the unsupervised dimensionality reduction by superpixel segmentation. (3) Since the regions obtained by superpixel segmentation have homogeneity, SuperPCA can extract potential low-dimensional features even under noise. (4) Although SuperPCA is an unsupervised method, it can achieve competitive performance when compared with supervised approaches. The resulting features are discriminative, compact, and noise resistant, leading to improved HSI classification performance. Experiments on three public datasets demonstrate that the SuperPCA model significantly outperforms the conventional PCA based dimensionality reduction baselines for HSI classification, and some state-of-the-art feature extraction approaches. The Matlab source code is available at \url{https://github.com/junjun-jiang/SuperPCA}.

\end{abstract}


\section{Introduction}
\label{sec:intro}
\IEEEPARstart{H}YPERSPECTRAL image (HSI) acquired by spaceborne or airborne sensors, such as AVIRIS, HyMap, HYDICE, and Hyperion, typically record material's hundreds of thousands of spectral wavelengths for each pixel in the image, which has opened new perspectives in many applications in remote sensing~\cite{fauvel2013advances, zhang2016deep, ma2018guided}. Since the subtle differences in ground covers can be captured by different spectral signatures, hyperspectral imagery is a well-suited technology for discriminating materials of interest. Although the rich spectral signatures can provide useful information for data analysis, the high dimensionality of HSI data presents some new challenges: (i) increasing the burden of data transmission and storage; (ii) leading to the curse of dimensionality problem~\cite{melgani2004classification} which will reduce the generalization capability of classifiers and deteriorate the classification performance, especially when the available labeled samples are limited. Owing to (i) the dense sampling of spectral wavelengths and (ii) the spectral reflectance of most materials changes only gradually over certain spectral bands, many contiguous bands are highly correlated and not all features (or spectral bands) are expected to contribute useful information for the data classification/analysis task at hand. As one of the typical method to alleviate this problem, dimensionality reduction is widely used as a preprocessing step to remove the highly correlated and redundant measurements in the original high-dimensional HSI spectral space and preserve essential information in a low-dimensional subspace. It has attracted increasing attentions in recent years. 

\begin{figure}[t]
\centering
\centerline{\includegraphics[width=8.80cm]{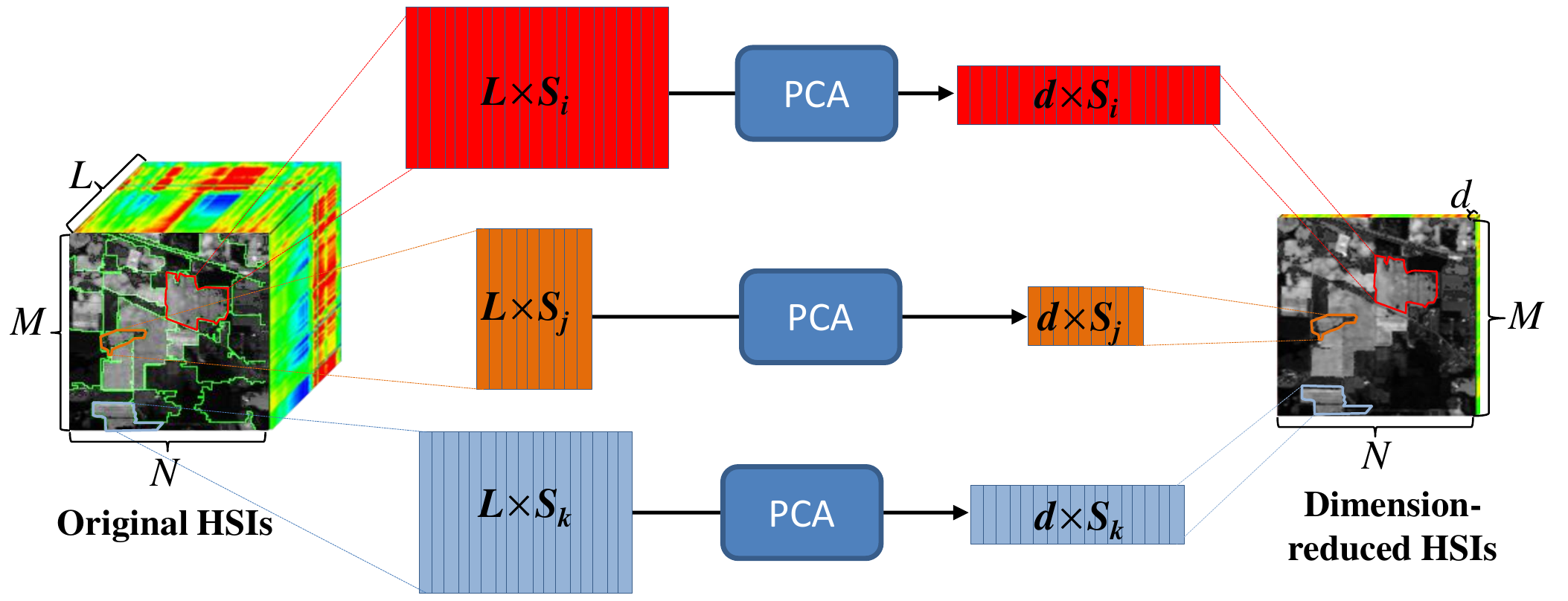}}
\caption{Schematic of \textcolor[rgb]{0.00,0.00,0.00}{SuperPCA} based dimensionality reduction for HSIs.}
\label{fig:SuperPCAfig}
\end{figure}

\begin{figure*}[!htpb]
\centering
\centerline{\includegraphics[width=14.00cm]{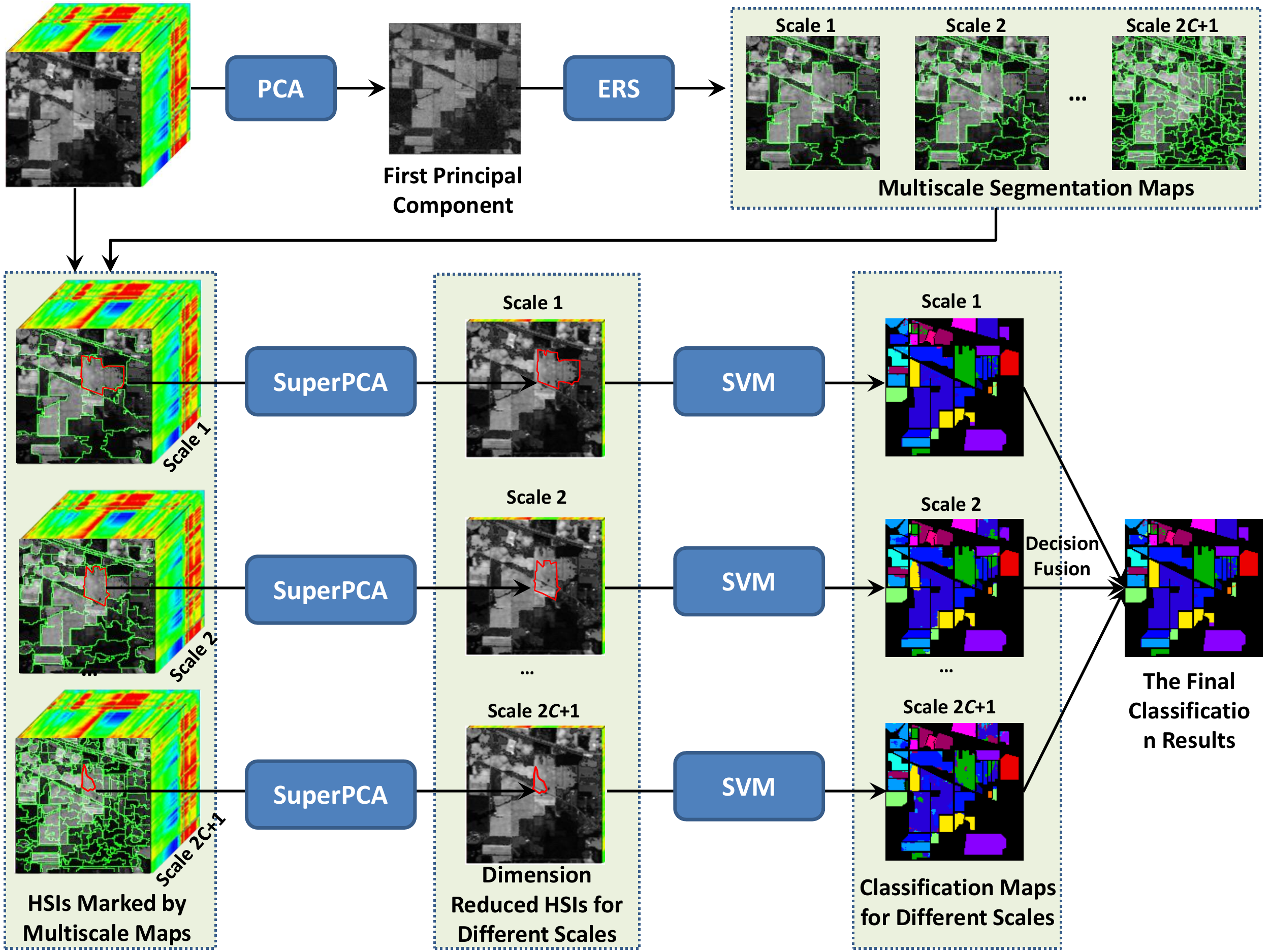}}
\caption{Outline of the proposed multiscale SuperPCA based HSI classification framework.}
\label{fig:framework}
\end{figure*}

Generally speaking, dimensionality reduction of HSI data can be divided into two categories: feature selection~\cite{du2008similarity, yang2011unsupervised, Wang2016Salient, jia2016novel} and feature extraction~\cite{bruce2002dimensionality, zhao2016spectral, sun2017sparse}. The former tends to select a small subset of the most representative bands from the original bands, whereas the latter aims to find an optimal transformation matrix to project the original \textcolor[rgb]{0.00,0.00,0.00}{high-dimensional} spectral \textcolor[rgb]{0.00,0.00,0.00}{features} into a \textcolor[rgb]{0.00,0.00,0.00}{low-dimensional} subspace. Feature selection can only select existing bands from HSIs, whereas feature extraction can use entire bands to generate more discriminative features. In \cite{zhang2016simultaneous}, a joint feature extraction and feature extraction method for HSI representation and classification has been developed. In this paper we mainly focus on employing feature extraction to reduce the feature \textcolor[rgb]{0.00,0.00,0.00}{dimensions} of HSIs. \textcolor[rgb]{0.00,0.00,0.00}{Based on whether or not the label information is used}, the feature extraction can be classified into unsupervised approaches and supervised approaches.

One of the most widely applied unsupervised \textcolor[rgb]{0.00,0.00,0.00}{dimensionality} reduction techniques in HSI analysis is the principal component analysis (PCA)~\cite{scholkopf1998nonlinear} and its variants~\cite{prasad2008limitations, hossain2011unsupervised, Laparra2015Dimensionality, ma2019infrared, zhao2016spectral}. Without any label information, PCA tends to find orthogonal transformations to maximize the total variance of the projected data. Different from preserving the largest data variance as in PCA, independent component analysis (ICA)~\cite{wang2006independent} tries to find the independent components by maximizing the statistical independence of the estimated components. Recently, some nonlinear methods based on manifold learning~\cite{roweis2000nonlinear, ma2017feature} have been used to compute the essential embedded low-dimensional space of observed high dimensional data~\cite{lunga2014manifold}, \emph{e.g.}, locally linear embedding (LLE)~\cite{bachmann2005exploiting, ma2010anomaly}, neighborhood preserving embedding (NPE)~\cite{he2005neighborhood}, locality preserving projection (LPP)~\cite{he2004locality}, and most recently proposed local pixel neighborhood preserving embedding (LPNPE)~\cite{zhou2015dimension}. Other efficient unsupervised feature extraction and learning methods also include intrinsic representation~\cite{Xu2016Intrinsic}, sub-feature learning~\cite{Slavkovikj2016Unsupervised}, and latent subclass learning~\cite{Wei2015Latent}. Supervised dimensionality reduction algorithms leverage the supervised information, \emph{i.e.}, the labels, to learn the dimensionality reduced feature space. The most representative works include Fisher's linear discriminant analysis (LDA)~\cite{prasad2008limitations} and Local Fisher discriminant analysis (LFDA)~\cite{li2012locality}.

Most of the above feature extrication methods use only spectral signature of each pixel and the dimensionality reduction models cannot directly use spatial information of HSIs, which has been proven to be very effective to improve the HSI representation and classification accuracy~\cite{fauvel2013advances,tarabalka2009spectral, fang2014spectral, wang2016spectral}. In~\cite{zhou2015dimension}, the spatial information is applied to spatial filtering (as a preprocessing) as well as modeling the spatial neighboring pixel correlations. Wen \emph{et al}. proposed to incorporate the spatial information, \emph{e.g.}, texture or morphological features, into the framework of orthogonal nonnegative matrix factorization~\cite{Wen2016TGRS}. The approach of \cite{Pu2014A} presents a novel spectral-spatial feature based similarity measurement which can be incorporated into existing dimensionality reduction methods including linear or nonlinear techniques. In \cite{ma2016spatial, jiang2017spatial}, spatial information is used to regularize the spectral representation.

\subsection{Motivation and Contributions}
Conventional methods usually learn a unified projection for HSI feature extraction~\cite{li2012locality, ly2014collaborative, ly2014sparse, li2016sparse}. However, different regions in an HSI may correspond to different objects, whose spectral features are diverse. Therefore, a reasonable way is to learn different projection matrices for different regions. Image segmentation can be seen as an exhaustive partitioning of the observed image into many different regions, and each of which is considered to be homogeneous~\cite{pal1993review}. These regions form a segmentation map that can be used as spatial structures for the spectral-spatial classification.

In this paper, we advocate a simple yet very effective unsupervised feature extraction method based on superpixelwise PCA, which is denoted as SuperPCA. It can learn the intrinsic low-dimensional features of different regions of the HSI data by performing PCA on each homogeneous region obtained by superpixel segmentation, as shown in Fig.~\ref{fig:SuperPCAfig}. An HSI is firstly divided into many homogeneous regions via superpixel segmentation, which are denoted by matrices whose columns are the spectral vectors of pixels. PCA is applied to these \textcolor[rgb]{0.00,0.00,0.00}{high-dimension} matrices to obtain the dimensionality reduced ones. Finally, we rearrange and combine all these low dimensional matrices to form the dimensionality reduced HSIs.

In an attempt to make full use of the spatial information contained in the HSI cube, we further develop a multiscale segmentation based SuperPCA model, namely MSuperPCA, which can effectively integrate multiscale spatial information to obtain the optimal classification result by decision fusion. Fig. \ref{fig:framework} demonstrates the schematic of our proposed multiscale SuperPCA method. We first apply \textcolor[rgb]{0.00,0.00,0.00}{entropy rate superpixel (ESR) to} obtain multiscale superpixel segmentations (by setting different superpixel numbers) based on the first principal component of the input HSIs. Then, for each scale, the \textcolor[rgb]{0.00,0.00,0.00}{proposed SuperPCA} based unsupervised dimensionality reduction method is used to obtain the dimensionality reduced HSIs. Based on the predictions of different scales through \textcolor[rgb]{0.00,0.00,0.00}{support vector machine} (SVM) classifier, we generate the final classification result via the majority voting decision fusion strategy.

To the best of our knowledge, this is the first time that a superpixelwise model is adopted for unsupervised dimensionality reduction and classification in hyperspectral imagery. Extensive experimental results demonstrate that, our method is not only simple and intuitive, but also achieves the most competitive HSI classification results as compared with the state-of-the-art dimensionality reduction based methods, including some recently proposed supervised feature extraction techniques. When the label information is limited (a small number of labeled training samples, e.g. 5 samples per class), our proposed SuperPCA and MSuperPCA methods obtain even better classification accuracies than the state-of-the-art supervised feature extraction techniques.

\subsection{Organization of This Paper}
The remainder of the paper is organized as follows. \textcolor[rgb]{0.00,0.00,0.00}{Section~\ref{sec:ers} firstly reviews and introduces the ESR superpixel segmentation algorithm.} Section~\ref{sec:SuperPCA} introduces notations and then explains the details of the proposed HSI classification approach based on SuperPCA and the multiscale extension of SuperPCA model. And then, we also give some analysis of the proposed SuperPCA algorithm. Section~\ref{sec:experiment} presents the experimental results and analysis. Finally, the concluding remarks are stated in Section~\ref{sec:conclusions}.

\section{Entropy Rate Super-pixel Segmentation (ERS)}
\label{sec:ers}
For a superpixel segmentation algorithm, it should have the following characteristics. Firstly, superpixels should adhere well to the object boundaries. Secondly, as a preprocessing process, superpixel segmentation should be of low computational complexity itself.

\textcolor[rgb]{0.00,0.00,0.00}{Recently, graph structure based segmentation approaches are widely used in superpixel segmentation~\cite{verdoja2015fast} and applications~\cite{Yan2013Hierarchical}.} A typical superpixel segmentation technique is the eigen-based solution to the normalized cuts (NCuts)~\cite{shi2000normalized}. However, it needs to construct a very large graph ($\mathscr{G}=(\textbf{V},\textbf{E})$) whose \textcolor[rgb]{0.00,0.00,0.00}{vertices} ($\textbf{V}$) are the pixels in the image to be segmented, the edge set ($\textbf{E}$) consists of the pairwise similarities by the weight function  $\textbf{s}:\textbf{E} \to {\mathbb{R}^ + } \cup \left\{ 0 \right\}$. Therefore, performing eigenvalue decomposition on such a large similarity matrix is very time consuming, which will take several minutes for segmenting an image of moderate size, \emph{e.g.}, around 500$\times$300 pixels. TurboPixel~\cite{levinshtein2009turbopixels} is an efficient alternative to achieve a similar regularity. However, it sacrifices fine image details and results in a low boundary recall. In \cite{liu2011entropy}, an ERS segmentation approach is proposed, and the graph is partitioned into a connected subgraph by choosing a subset of edges $\textbf{A} \subseteq \textbf{E}$ such that the resulting graph $\mathscr{G}^{'}=(\textbf{V}, \textbf{A})$ consists of smaller connected components/subgraphs. In the objective function of ERS, it incorporates an entropy rate term $H(\textbf{A})$ and a balancing term $B(\textbf{A})$ to optimize the superpixel segmentation:
\begin{equation}\label{eq:ers}
\textbf {A}^*\mathop {{\rm{ = argmax}}}\limits_\textbf{A} {\rm{Tr}}\left\{ {H(\textbf{A}) + \alpha B(\textbf{A})} \right\},{\kern 1pt} {\kern 1pt} {\kern 1pt} {\kern 1pt} {\kern 1pt} {\kern 1pt} {\kern 1pt} {\kern 1pt}  s.t.{\kern 1pt} {\kern 1pt} {\kern 1pt} {\kern 1pt} {\kern 1pt} \textbf{A} \subseteq \textbf{E}.
\end{equation}
Here, the first term favors the formation of homogeneous and compact clusters, while the second term can be used to encourage the cluster with similar sizes. $\alpha$ is used to balance the contributions of the entropy rate term $H(\textbf{A})$ and the balancing term $B(\textbf{A})$. As described in \cite{nemhauser1978analysis}, a greedy algorithm effectively solves the optimization problem in (\ref{eq:ers}). This method is highly efficient, which only takes about 2.5 seconds to segment an image of size 500$\times$300 pixels.

\section{Superpixelwise Principal Component Analysis (SuperPCA)}
\label{sec:SuperPCA}
An HSI cube ${\textbf{X}} \in {\mathbb{R}^{M \times N \times L}}$ is made up with hundreds of nearly contiguous spectral bands, with high (5-10 nm) spectral resolution, from the visible to infrared spectrum for each image pixel. Here, $M$, $N$ and $L$ are the number of image rows, columns and sampled wavelengths, respectively. We can reshape the 3D cube to a 2D matrix, ${\textbf{X}=}\left[ {{{\textbf{x}}_1},{{\textbf{x}}_2}, \cdots {\rm{,}} {{\textbf{x}}_P}} \right] \in {\mathbb{R}^{L \times P}}$  ($P = MN$), in which each column represents one pixel vector that reflects the energy spectrum of the materials within the spatial area covered by the pixel.

Denote ${\textbf{x}_i} \in {\mathbb{R}^L} (1 \le i \le P)$ the $i$-th pixel vector of the observed HSI cube $\textbf{X} \in {\mathbb{R}^{L \times P}}$,
\begin{equation}\label{eq:j1}
{{\textbf{x}}_i} = {\left[ {{x_{i1}},{x_{i2}}, \cdots ,{x_{iL}}} \right]^T}.
\end{equation}
PCA performs the dimensionality reduction by computing the low-dimensional representation that maximizes data variance in the dimensionality reduced space. Specifically, it finds a linear mapping from the original $L$-dimensional space ${\textbf{X}} \in {\mathbb{R}^{L \times P}}$ to a low $d$-dimensional space ${\textbf{Y = }}\left[ {{{\textbf{y}}_1}{\rm{,}}{{\textbf{y}}_2}{\rm{,}} \cdots {\rm{,}}{{\textbf{y}}_P}} \right] \in {\mathbb{R}^{d \times P}}$, $d < L$. Without loss of generality, we denote the transformation matrix by $\textbf{W}$. That is, ${{\textbf{y}}_i} = {\textbf{W}^T}{{\textbf{x}}_i}$. Mathematically, it aims at finding the linear transformation matrix by solving the following objective function,
\begin{equation}\label{eq:j1}
\textbf {W}^*\mathop {{\rm{ = argmax}}}\limits_{{\textbf{W}^T}\textbf{W} = I} {\rm{Tr}}\left( {{\textbf{W}^T}{\rm{Cov}}\left( {\textbf{X}} \right){\textbf{W}}} \right),
\end{equation}
where ${\rm{Cov}}\left( {\textbf{X}} \right)$ stands for the covariance matrix of the data set $\textbf{X}$, and Tr($\textbf{X}$) denotes the trace of an $n$-by-$n$ square matrix $\textbf{X}$.

\begin{figure}[!htbp]
\centering
\centerline{\includegraphics[width=5.5cm]{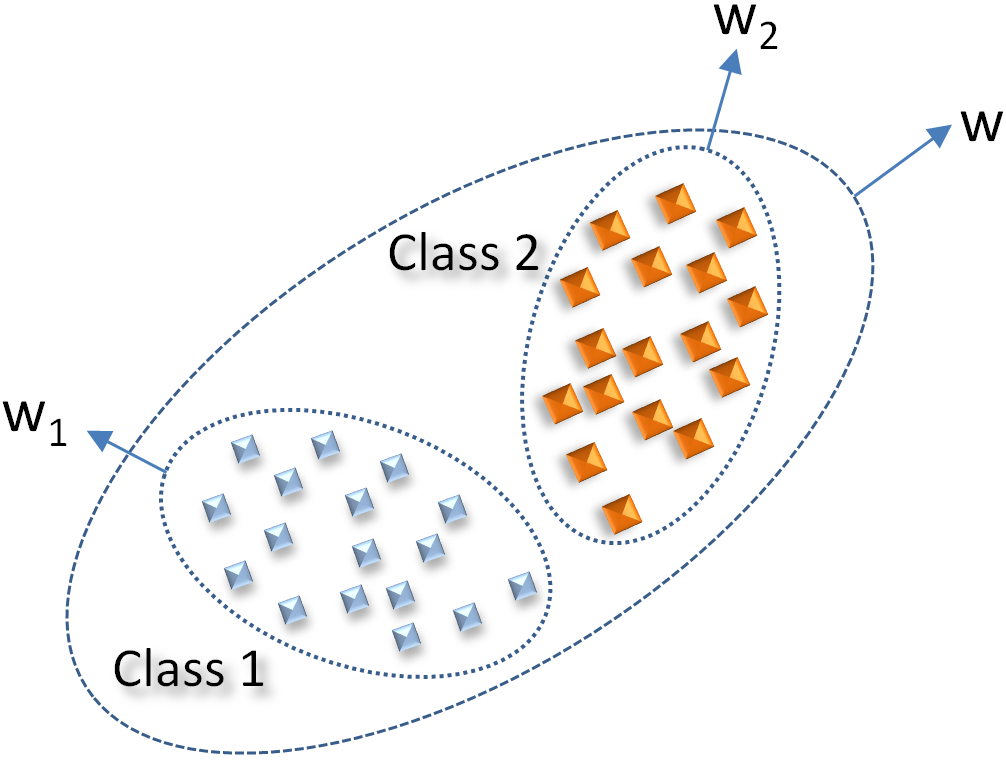}}
\caption{The principal projection directions of global PCA and class specific PCA.}
\label{fig:twoclasses}
\end{figure}

\begin{figure}[!t]
\centering
\centerline{\includegraphics[width=8.80cm]{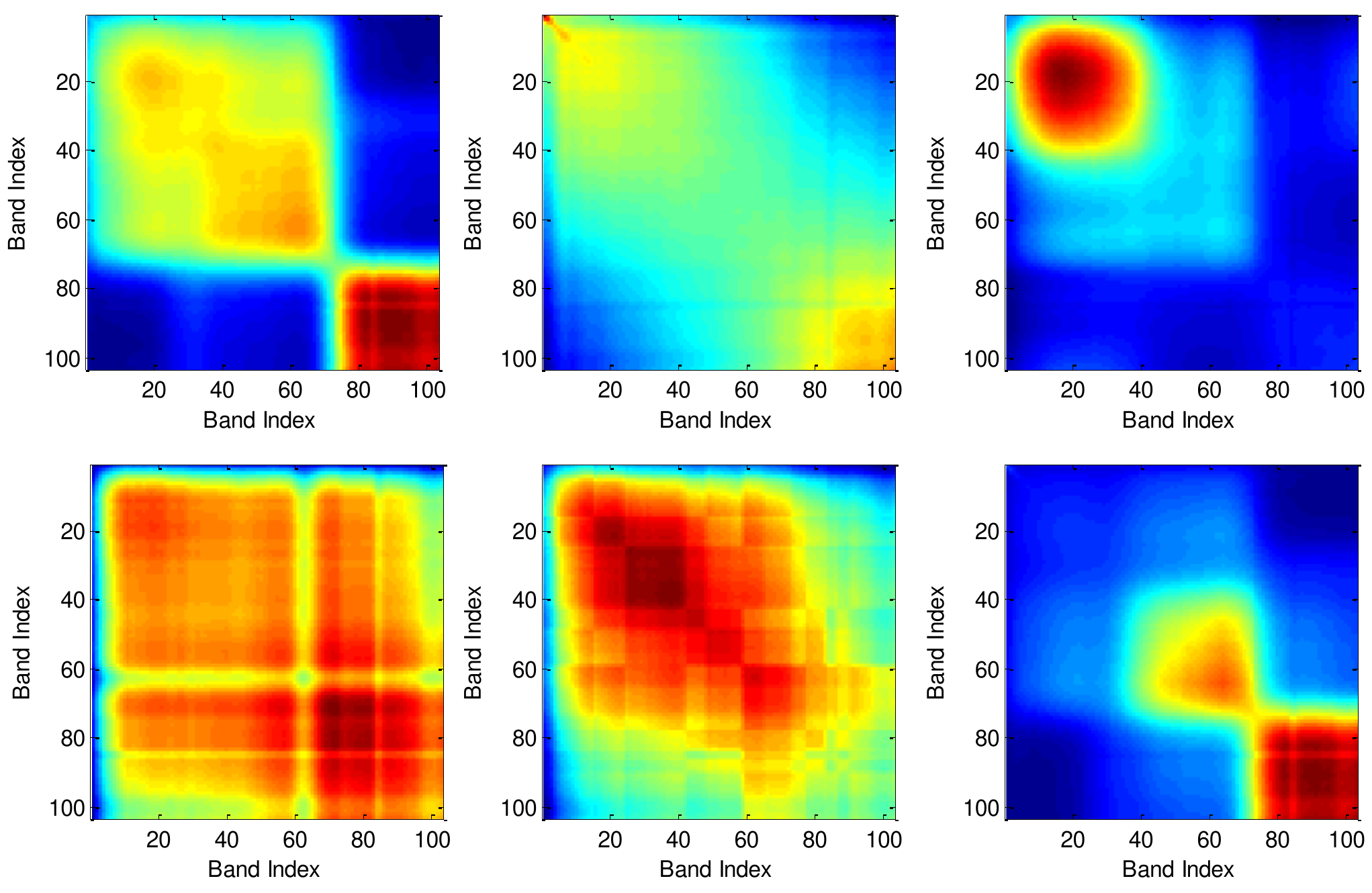}}
\vspace{-0.20cm}
\caption{Visualization of the correlation matrices of spectral bands of the entire University of Pavia image (the top left subfigure) and different homogeneous regions (the rest subfigures).}
\label{fig:corrMatrix}
\end{figure}

Owing to its simplicity, effectiveness, and robustness to noise, PCA has been widely used as a preprocessing step of many HSI based applications. However, in an HSI, there are many homogeneous regions. Within each region, pixels are more likely to be the same class~\cite{li2015efficient, fang2015spectral,li2013classification,zhang2018one}. The global PCA approach considers the entire data space (composed of all the pixel vectors of the HSI cube), and tries to find the best transformation vector for this space. It may ignore the differences of homogeneous regions. As illustrated by a toy example (Fig. \ref{fig:twoclasses}), we suppose that the data space is formed by class 1 (marked with blue squares) and class 2 (marked with orange squares), which could possibly represent distributions of samples from two different homogeneous regions of HSIs. We can obviously see that the transformation vectors $\textbf{w}_1$ and $\textbf{w}_2$ for class 1 and class 2 are significantly different, and they are also different from the transformation vector $\textbf{w}$ generated for the entire data space. \textcolor[rgb]{0.00,0.00,0.00}{As shown in Fig.~\ref{fig:corrMatrix}, we plot the correlation matrices of spectral bands of the entire University of Pavia image as well as some typical homogeneous
regions. From this figure, we can learn that the correlation matrices are variant. Therefore, different regions will have varying transformation vectors (see Eq.~(\ref{fig:twoclasses}))}.

\subsection{Generation of Homogeneous Regions}
Inspired by the above observation, in this paper we propose a divide-and-conquer strategy to perform unsupervised feature extraction based on PCA for each homogeneous region. By extracting the same number of principal components (PCs) for each homogeneous region, we can combine them to form the dimensionality reduced HSIs (Fig. \ref{fig:framework}). In the following, we will introduce the construction of homogeneous regions using superpixel segmentation, which can exhaustively partition the image into many homogeneous regions.

\textcolor[rgb]{0.00,0.00,0.00}{As in many superpixel segmentation based hyperspectral image classification and restoration methods \cite{li2015efficient, fang2015spectral, zhang2017multiscale, Fan2017Hyperspectral}, we adopt ERS due to its promising performance in both efficiency and efficacy. Other state-of-the-art methods such as simple linear iterative clustering (SLIC) \cite{Achanta2012SLIC} can also be used to replace the ERS.} Specially, we first obtain the first principal component of HSIs, $I_f$, capturing the major information of HSIs. This further reduces the computational cost for superpixel segmentation. And then, we perform ESR on $I_f$ to obtain the superpixel segmentation,
\begin{equation}\label{eq:segmentation}
I_f = \bigcup\limits_k^S {{\mathscr{X}_k}} ,{\kern 1pt} {\kern 1pt} {\kern 1pt} {\kern 1pt} {\kern 1pt} s.t.{\kern 1pt} {\kern 1pt} {\kern 1pt} {\mathscr{X}_k} \cap {\mathscr{X}_g} = \emptyset ,{\kern 1pt} {\kern 1pt} {\kern 1pt} (k \ne g),
\end{equation}
where $S$ denotes the number of superpixels, and $\mathscr{X}_k$ is the $k$-th superpixel.

\subsection{Multiscale Extension of SuperPCA}
By segmenting the HSIs to superpixels, it will be beneficial to exploit rich spatial information about the land surface~\cite{zhang2017multiscale,fang2014spectral}. However, how to select an optimal value for the number of superpixels is a very challenging problem in actual applications~\cite{liu2011entropy}. When the superpixels are too large (by setting a small superpixel number), the resultant under-segmentation can lead to ambiguity-labeled boundary superpixels that require further segmentation. When superpixels are too small (by setting a large superpixel number), the features computed from the over-segmented regions may become less distinctive, making it more difficult to infer correct labels. In addition, as reported in \cite{coburn2004multiscale}, there is no single region size that would adequately characterize the spatial information of HSIs. Inspired by the classifier and decision fusion techniques \cite{prasad2008decision, ding2017local, ma2016infrared}, in this paper we propose the multiscale segmentation strategy to enhance the performance of single scale SuperPCA based method, thus alleviating above-mentioned problem. More specifically, the principal component image $I_f$ (the first  principal component of HSIs) is segmented into $2C+1$ scales. The number of superpixel of the $c$-th scale is $S_c$,
\begin{equation}\label{eq:multiseg}
{S_c} = {(\sqrt 2) ^c}{S_f},{\kern 1pt} {\kern 1pt} {\kern 1pt} {\kern 1pt} c = 0, \pm 1, \pm 2, \cdots , \pm C,
\end{equation}
where $S_f$ is the fundamental superpixel number and is set empirically. Since the value of $S_c$ may not be an integer number in $\{1, 2, ... ,P\}$, we reset it as $S_c = min(max(1; round(S_c));P)$. Here, $P$ is the number of total pixels in the HSIs.

By taking advantage of the multiscale superpixels, the decision fusion strategy can boost the classification accuracy, especially in conflicting situations. Specifically, we fuse the label information of each test pixel predicted by different multiscale superpixels. That is, given that the fundamental image $I_f$ is segmented to $2C+1$ scales, and there will be $2C+1$ different classification results for an HSI. Then, we can aggregate the results through an effective decision fusion strategy. In this paper, we leverage the majority voting (MV) based decision fusion strategy due to its \textcolor[rgb]{0.00,0.00,0.00}{insensitivity} to inaccurate estimates of posterior probabilities:
\begin{equation}\label{eq:majorityvoting}
l = \mathop {\arg \max }\limits_{i \in \{ 1,2,...,G\} } N(i),{\kern 5pt} N(i) = \sum\limits_{j = 1}^{2C+1} \alpha_j{I({l_j} = i)},
\end{equation}
where $l$ is the class label from one
of the $G$ possible classes for the test pixel, $j$ is the classifier index, $N(i)$ represents the number of times that class $i$ is predicted in the bank of classifiers, and $I$ denotes the indicator function. In Eq. (\ref{eq:majorityvoting}), $\alpha_j$ denotes the voting strength of the $j$-th classifier. One possible way of performing this adaptive voting mechanism is to weigh a classifier's vote based on its confidence score, which can be learned from training data. In this paper, we directly use the equal voting strength, $\alpha_j  = \frac{1}{{2{\rm{C}} + 1}}{\kern 4pt}  (j = 1,2,...,2C+1)$.

Fig.~\ref{fig:framework} shows the framework of the proposed multiscale SuperPCA method for HSI classification. We firstly obtain the first principal component of the input HSIs. Then, it is segmented to multiple scales based on the ESR algorithm~\cite{liu2011entropy} with different superpixel numbers. For each scale, we perform PCA dimensionality reduction on each homogeneous region and combine all regions to form the \textcolor[rgb]{0.00,0.00,0.00}{dimension-reduced} HSIs. Lastly, we apply SVM classification to each \textcolor[rgb]{0.00,0.00,0.00}{dimension-reduced} HSIs and fuse the classification results by majority voting to predict the final labels for testing samples.

\begin{figure}[t]
\centering
\centerline{\includegraphics[width=8.9cm]{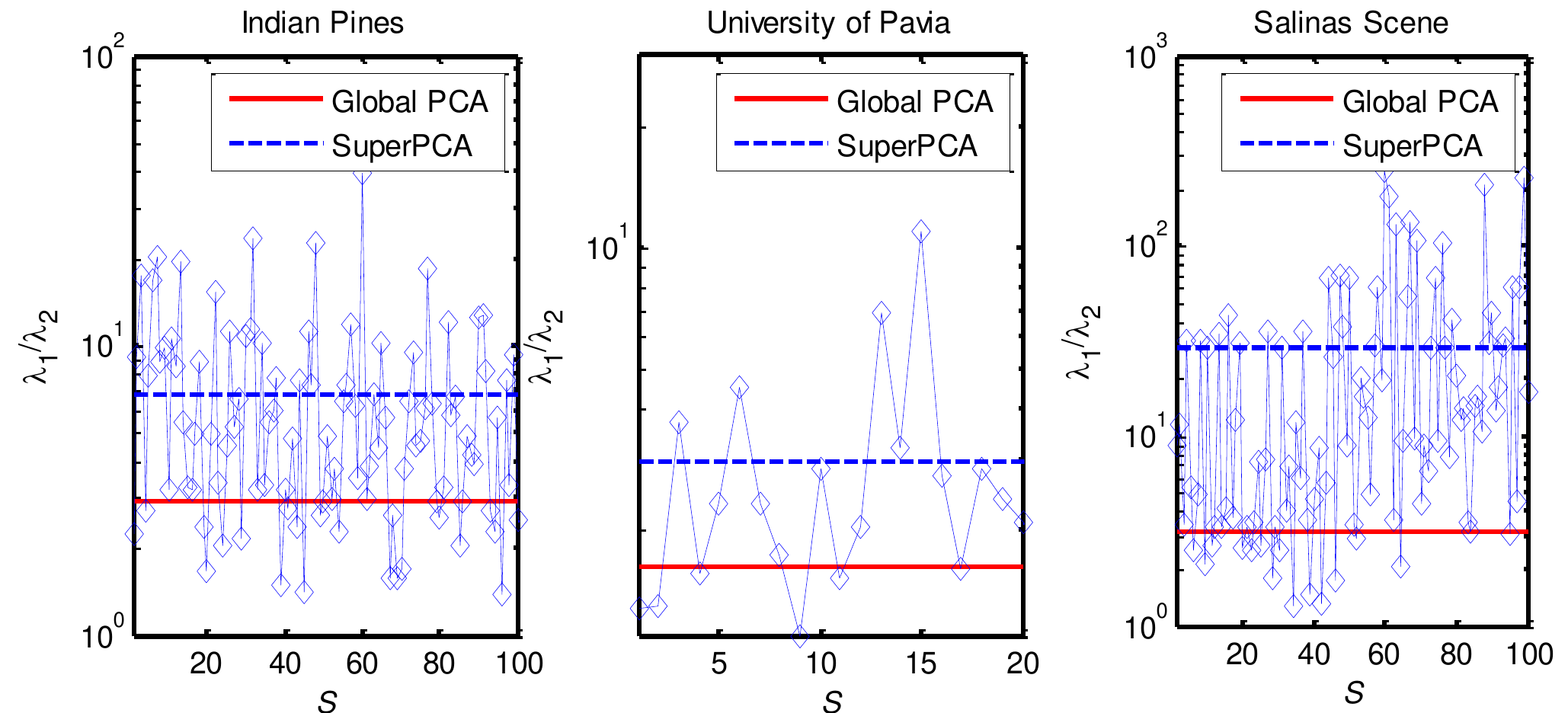}}
\vspace{-0.20cm}
\caption{The ratio between the first and second eigenvalues ($\lambda_1/\lambda_2$). The red lines are the ratios of the global PCA method, while the blue plots are the ratios of all the homogeneous regions based on the proposed SuperPCA method when the number of superpixel is set to the optimal value, $S=100$, $S=20$, and $S=100$, for Indian Pines, University of Pavia, and Salinas Scene, respectively. The blue horizontal line represents the average ratio of all the homogeneous regions. For the convenience of observation, we use a logarithmic scale for the values of ratios.}
\label{fig:ratios}
\end{figure}

\subsection{Analysis of the Proposed SuperPCA}
\emph{Remark 1}.
\textcolor[rgb]{0.00,0.00,0.00}{Through superpixel segmentation, we can obtain different homogeneous regions, in which pixels are more likely to fall in the same class~\cite{li2015efficient, fang2015spectral,li2013classification}. By dividing the global HSIs to some small regions, it becomes easier to find the intrinsic projection directions.} Fig. \ref{fig:ratios} shows the ratios between the first and second eigenvalues of PCA (global based) and the proposed SuperPCA on Indian Pines, University of Pavia, and Salinas Scene HSI datasets (for more detailed information about the datasets and the parameter setting of the number of superpixel $S$, please refer to the experimental section). Obviously, the larger the ratio, the more representative and discriminant the primary projected features are. By segmenting the HSIs to different homogeneous regions, SuperPCA gains larger ratio than conventional global PCA method (see the blue and red horizontal lines). It is worth noting that larger $S$ results in smaller homogeneous regions, and each of which has a better consistency. However, it does not necessarily lead to better classification performance. This is because, when the homogeneous region (superpixel) is too small, there will be few data samples in each superpixel, which may cause instability for PCA. From the experimental analysis, it is clear that the divide-and-conquer strategy of unsupervised feature extraction based on SuperPCA can significantly increase the eccentricity in the direction of the first eigenvector. This further corroborates our claim that a homogeneous region based PCA will be more effective in preserving the essential data information in a low dimensional space.

\emph{Remark 2}.
\textcolor[rgb]{0.00,0.00,0.00}{There are currently a number of region-based PCA methods for feature extraction or other related applications. For example, in region-based PCA face recognition~\cite{jiang2017patch, zhao2009part, gao2017semi}, they divide the whole face image into small patches, and then use PCA to extract the local features that cannot be captured by traditional global face based PCA algorithm; in region-based PCA image denosing~\cite{deledalle2011image}, they first divide the whole face image into small patches, and then stack similar noisy patches and apply PCA to exploit these consistency structure among similar patches (thus removing the noise). However, when we directly apply the regular patch based PCA algorithm to hyperspectral images, it cannot fully exploit the rich spatial information contained in HSIs. To this end, we propose a novel region-based PCA through superpixel segmentation strategy.} Table \ref{tab:segPCA} shows the average overall classification accuracies of three divide-and-conquer strategies\footnote{The differences of these three strategies lie in their dividing strategies. In ClusterPCA, all the pixels are clustered by $K$-means, and then PCA is applied to each cluster to obtain the dimensionality reduced features. SquarePCA directly performs PCA dimensionality reduction on the squared patches of HSIs.}, Clustering dependent PCA (ClusterPCA for short), Square patch dependent PCA (SquarePCA for short), and the proposed SuperPCA, with different training sample numbers on the Indian Pines dataset. In addition, the Global PCA method is used as a baseline for comparison. Without loss of generality, we only conduct experiments on this dataset and similar conclusions can be found on the other two datasets.

\setlength{\tabcolsep}{2.35pt}
\begin{table}[t]
\scriptsize
\centering
\caption{Classification results (in terms of OA) of three divide-and-conquer strategies on the Indian Pines dataset using SVM and NN classifiers.}
\label{tab:segPCA}
\vspace{1pt}
\begin{tabular}{|c|c|c|c|c|c|}
\hline
Noise & T.N.s/C	&	Global PCA	&	ClusterPCA	&	SquarePCA	&	SuperPCA	\\
\hline
\hline
\multirow{4}{*}{$\sigma=0$}&5	&	46.37, 46.94	&	46.37, 46.94	&	67.32, 65.64	&	77.34, 75.85	\\
&10	&	55.72, 52.06	&	55.72, 51.83	&	77.59, 76.89	&	85.76, 83.79	\\
&20	&	62.97, 56.88	&	62.97, 56.65	&	84.32, 83.97	&	92.87, 91.94	\\
&30	&	67.27, 59.50	&	67.27, 59.37	&	87.36, 87.02	&	94.62, 93.78	\\
\hline
\hline
\multirow{4}{*}{$\sigma=10$}&5	&	35.25, 36.17	&	37.08, 36.09	&	64.68, 63.49	&	74.26, 74.20	\\
&10	&	38.63, 37.68	&	39.76, 39.08	&	75.95, 75.14	&	82.52, 82.18	\\
&20	&	44.40, 39.65	&	44.40, 41.19	&	81.71, 81.33	&	90.42, 89.05	\\
&30	&	45.51, 41.13	&	45.51, 42.04	&	84.00, 83.86	&	93.36, 90.84	\\
\hline
\end{tabular}
\end{table}

\begin{table*}[t]
\scriptsize
      \caption{Number of samples in the Indian Pines, University of Pavia, and Salinas Scene images}
      \centering
        \begin{tabular}[ht]{ |c| c || c | c  || c| c| }
\hline
\multicolumn{2}{|c||}{Indian Pines}  &\multicolumn{2}{c||}{University of Pavia}  &\multicolumn{2}{c|}{Salinas Scene} \\
\hline
Class Names	&	Numbers	&	Class Names	&	Numbers	&	Class Names	&	Numbers	\\
\hline
\hline
 \multicolumn{1}{>{\columncolor{indian1}}c}{Alfalfa}	&	46	&	 \multicolumn{1}{>{\columncolor{paviaU1}}c}{Asphalt}	&	6631	&	 \multicolumn{1}{>{\columncolor{indian1}}c}{Brocoli\_green\_weeds\_1}	&	2009	\\
 \multicolumn{1}{>{\columncolor{indian2}}c}{Corn-notill}	&	1428	&	\multicolumn{1}{>{\columncolor{paviaU2}}c}{Bare soil}	&	18649	&	 \multicolumn{1}{>{\columncolor{indian2}}c}{Brocoli\_green\_weeds\_2} 	&	3726	\\
 \multicolumn{1}{>{\columncolor{indian3}}c}{Corn-mintill}	&	830	&	\multicolumn{1}{>{\columncolor{paviaU3}}c}{Bitumen}	&	2099	&	 \multicolumn{1}{>{\columncolor{indian3}}c}{Fallow} 	&	1976	\\
 \multicolumn{1}{>{\columncolor{indian4}}c}{Corn}	&	237	&	\multicolumn{1}{>{\columncolor{paviaU4}}c}{Bricks}	&	3064	&	 \multicolumn{1}{>{\columncolor{indian4}}c}{Fallow\_rough\_plow} 	&	1394	\\
 \multicolumn{1}{>{\columncolor{indian5}}c}{Grass-pasture}	&	483	&	\multicolumn{1}{>{\columncolor{paviaU5}}c}{Gravel}	&	1345	&	 \multicolumn{1}{>{\columncolor{indian5}}c}{Fallow\_smooth} 	&	2678	\\
 \multicolumn{1}{>{\columncolor{indian6}}c}{Grass-trees}	&	730	&	\multicolumn{1}{>{\columncolor{paviaU6}}c}{Meadows}	&	5029	&	 \multicolumn{1}{>{\columncolor{indian6}}c}{Stubble} 	&	3959	\\
 \multicolumn{1}{>{\columncolor{indian7}}c}{Grass-pasture-mowed}	&	28	&	\multicolumn{1}{>{\columncolor{paviaU7}}c}{Metal sheets}	&	1330	&	 \multicolumn{1}{>{\columncolor{indian7}}c}{Celery} 	&	3579	\\
 \multicolumn{1}{>{\columncolor{indian8}}c}{Hay-windrowed}	&	478	&	\multicolumn{1}{>{\columncolor{paviaU8}}c}{Shadows}	&	3682	&	 \multicolumn{1}{>{\columncolor{indian8}}c}{Grapes\_untrained} 	&	11271	\\
 \multicolumn{1}{>{\columncolor{indian9}}c}{Oats}	&	20	&		\multicolumn{1}{>{\columncolor{paviaU9}}c}{Trees} &	947	&	 \multicolumn{1}{>{\columncolor{indian9}}c}{Soil\_vinyard\_develop} 	&	6203	\\
 \multicolumn{1}{>{\columncolor{indian10}}c}{Soybean-notill}	&	972	&		&		&	 \multicolumn{1}{>{\columncolor{indian10}}c}{Corn\_senesced\_green\_weeds} 	&	3278	\\
 \multicolumn{1}{>{\columncolor{indian11}}c}{Soybean-mintill} 	&	2455	&		&		&	 \multicolumn{1}{>{\columncolor{indian11}}c}{Lettuce\_romaine\_4wk} 	&	1068	\\
 \multicolumn{1}{>{\columncolor{indian12}}c}{Soybean-clean}	&	593	&		&		&	 \multicolumn{1}{>{\columncolor{indian12}}c}{Lettuce\_romaine\_5wk} 	&	1927	\\
 \multicolumn{1}{>{\columncolor{indian13}}c}{Wheat}	&	205	&		&		&	 \multicolumn{1}{>{\columncolor{indian13}}c}{Lettuce\_romaine\_6wk} 	&	916	\\
 \multicolumn{1}{>{\columncolor{indian14}}c}{Woods}	&	1265	&		&		&	 \multicolumn{1}{>{\columncolor{indian14}}c}{Lettuce\_romaine\_7wk} 	&	1070	\\
 \multicolumn{1}{>{\columncolor{indian15}}c}{Buildings-Grass-Trees-Drives}	&	386	&		&		&	 \multicolumn{1}{>{\columncolor{indian15}}c}{Vinyard\_untrained} 	&	7268	\\
 \multicolumn{1}{>{\columncolor{indian16}}c}{Stone-Steel-Towers} 	&	93	&		&		&	 \multicolumn{1}{>{\columncolor{indian16}}c}{Vinyard\_vertical\_trellis}	&	1807	\\
\hline
Total Number&10249&Total Number&42776&Total Number&54129\\
\hline
\end{tabular}
\label{table:Three_sample}
\end{table*}

\textcolor[rgb]{0.00,0.00,0.00}{It is should be noted that we use two different classifiers to conduct the classification, i.e., SVM and nearest neighbor (NN). For each result in the bracket, the left is based on the SVM classifier while the right is based on the NN classifier, respectively.} To evaluate the performance, we randomly choose $N=5, 10, 20, 30$ samples from each class to form the training set\footnote{At a maximum half of the total samples in Grass-pasture-mowed and Oats classes, which have relatively small sample sizes, are chosen.}, and the rest \textcolor[rgb]{0.00,0.00,0.00}{of the} samples for testing. \textcolor[rgb]{0.00,0.00,0.00}{Due to space limitation, we use ``T.N.s/C'' to denote training numbers in each class in the table.} In comparison to ClusterPCA and SquarePCA, the proposed SuperPCA method is more efficient. Global PCA and ClusterPCA have the similar results, which indicates that the preprocessing of clustering is invalid. This is because ClusterPCA does not use the spatial information, and considers each pixel as an isolated data sample. In contrast, SquarePCA and SuperPCA leverage the spatial information inside a square patch or a superpixel region, thus leading to better performance. Such advantage becomes more obvious in the case of noise presence. To demonstrate this, we add additive white Gaussian noise (AWGN) with the variance of $\sigma=10$ to the original HSIs. Please refer to the third block in Table \ref{tab:segPCA}. The performance of ClusterPCA drops drastically when adding noise, while SquarePCA and SuperPCA methods are less affected by the noise. \textcolor[rgb]{0.00,0.00,0.00}{For example, when the noise level is $\sigma=10$ and the number of training samples per class is 30, the classification accuracy of ClusterPCA is less than 50\%, while SquarePCA and SuperPCA can go beyond 80\%. Our SuperPCA method even reaches 93.36\% (for SVM classifier) and 90.48\% (for NN classifier)}. In all cases, our method achieves the best performance. In comparison to the SquarePCA method, which also takes into account the spatial information, our proposed SuperPCA method also yields significant performance gains, with an average of 8\% increase \textcolor[rgb]{0.00,0.00,0.00}{no matter what kind of classifier is used}. \textcolor[rgb]{0.00,0.00,0.00}{We attributes this superiority of SuperPCA over SquarePCA to that pixels in a superpixel are much more like to be the same class than those in a regular patch, and our method can exploit the spatial information more effectively. In summary,} clearly demonstrates the robustness of SuperPCA to noise in HSIs for image classification.

\section{Experimental Results and analysis}
\label{sec:experiment}
In this section, we first introduce the three HSI datasets used in our experiments. Then, we assess the impact of the number of superpixels and the reduced dimension on the classification performance using SuperPCA. The comparison results with the state-of-the-art \textcolor[rgb]{0.00,0.00,0.00}{dimensionality} reduction approaches are presented.

\subsection{Datasets and Experimental Procedure}
In order to evaluate the proposed SuperPCA method, we use three publicly available HSI datasets\footnote{\url{http://www.ehu.eus/ccwintco/index.php/Hyperspectral_Remote_Sensing_Scenes}}.
\begin{enumerate}
  \item The first HSI dataset is the \emph{Indian Pine}, which is acquired by the AVIRIS sensor in June 1992. The scene is with 145$\times$145 pixels and 220 bands in the 0.4-2.45 ¦Ìm region covering the agricultural fields with regular geometry. In this paper, 20 low SNR bands are removed and a total of 200 bands are used for classification. \textcolor[rgb]{0.00,0.00,0.00}{It contains 16 different land-covers, and approximately 10249 labeled pixels are from the ground-truth map}.
   \item The second HSI dataset is the \emph{University of Pavia}, which contains a spatial coverage of 610$\times$340 pixels and is collected by the ROSIS sensor under the HySens project managed by DLR (the German Aerospace Agency). It generates 115 spectral bands, of which 12 noisy and  water-bands are removed. It has a spectral coverage from 0.43-0.86 $\mu$m and a spatial resolution of 1.3 m. Approximately 42776 labeled pixels with nine classes are from the ground truth map.
  \item The third HSI dataset is the \emph{Salinas Scene}, collected by the 224-band AVIRIS sensor over Salinas Valley, California, capturing an area over Salinas Valley, CA, USA. It generates 512$\times$217 pixels and 204 bands over 0.4-2.5 $\mu$m with spatial resolution of 3.7 m, of which 20 water absorption bands are removed before classification. \textcolor[rgb]{0.00,0.00,0.00}{In this image, there are approximately 54129 labeled pixels with 16 classes sampled from the ground truth map}.
\end{enumerate}

For the three datasets, the training and testing samples are randomly selected from the available ground truth maps. The class-specific numbers of labeled samples are shown in Table~\ref{table:Three_sample}.
To evaluate the performance of our proposed SuperPCA algorithm, we randomly choose $T = 5, 10, 20, 30$ samples from each class to build the training set, leaving the rest samples to form the testing set. For some classes, \emph{e.g.}, Grass-pasture-mowed and Oats in the Indian Pines image, which have a few labeled samples, we only select a maximum half of the total samples in them. To avoid any bias, all the experiments are repeated 10 times, and we report the average classification accuracy.

\textcolor[rgb]{0.00,0.00,0.00}{We compare the proposed methods with two baseline methods (raw spectral features based and PCA method), as well as the state-of-the-art \textcolor[rgb]{0.00,0.00,0.00}{dimensionality} reduction approaches, including
five unsupervised feature extraction methods (PCA~\cite{scholkopf1998nonlinear}, ICA \cite{wang2006independent}, LPP~\cite{he2004locality}, NPE~\cite{he2005neighborhood} and LPNPE~\cite{zhou2015dimension}), and two supervised feature extraction methods (LDA~\cite{prasad2008limitations} and LFDA~\cite{li2012locality})}. Similar to many previous representative works~\cite{tarabalka2009spectral, tarabalka2010svm, camps2006composite}, three measurements, overall accuracy (OA), average accuracy (AA) and Kappa, are used to evaluate the performance of different \textcolor[rgb]{0.00,0.00,0.00}{dimensionality} reduction algorithms for HSI classification. Similar to [25], all above-mentioned feature extraction methods are performed on the filtered data using a 5$\times$5 weighted mean filter, and then \textcolor[rgb]{0.00,0.00,0.00}{SVM classifier is applied to filtered data. It is worth noting that for all the comparison methods, they all go through these two procedures. Firstly, they extract the features of the input HSIs with an unsupervised or supervised manner, and then the supervised SVM classifier is adopted to test their classification performances.}

\begin{figure}[!htbp]
\centering
\centerline{\includegraphics[width=8.80cm]{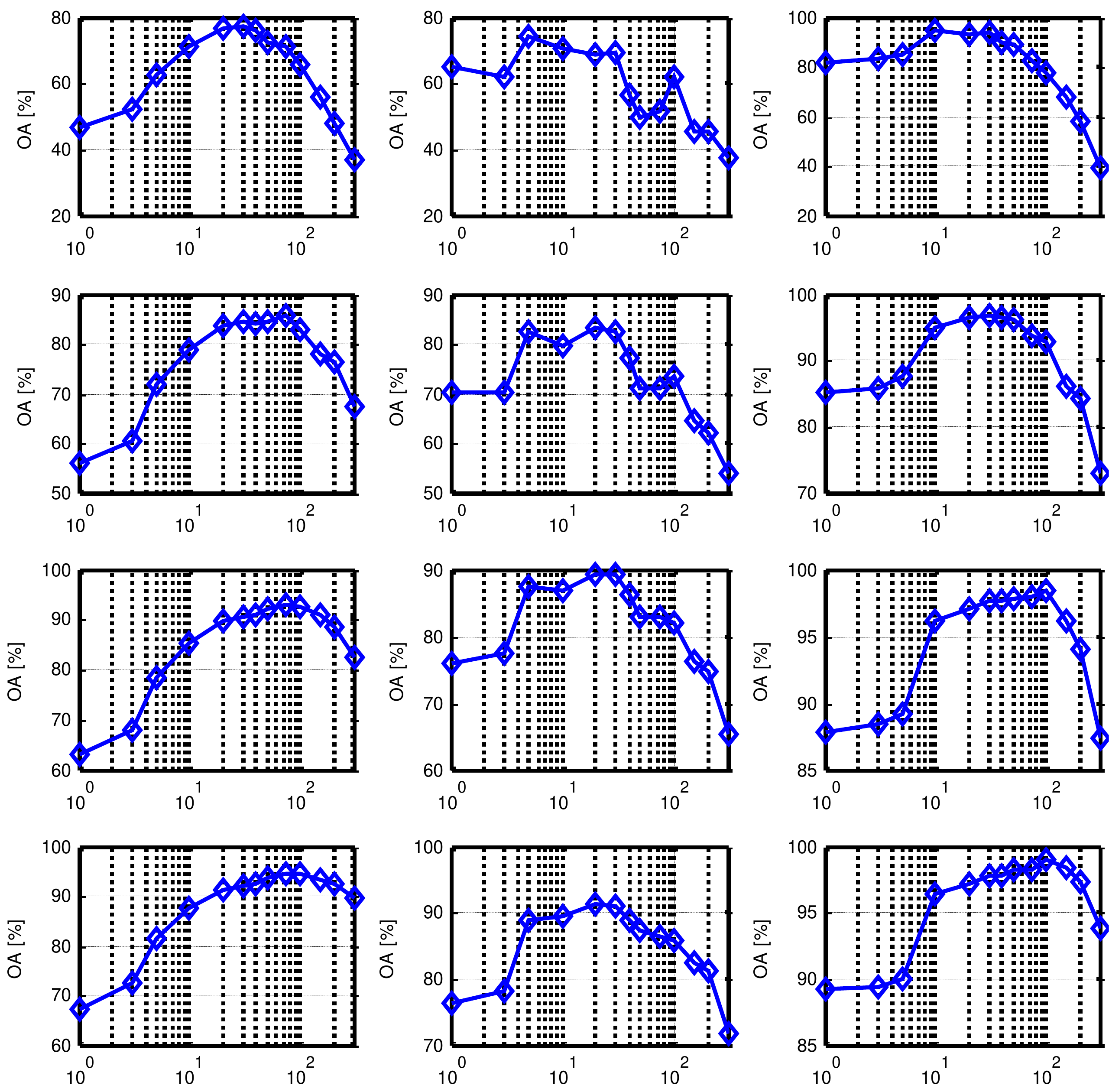}}
\vspace{-0.20cm}
\caption{The influences of the number of superpixels on the overall classification accuracy (\%) of the proposed SuperPCA method for Indian Pines (first column), University of Pavia (second column), and Salinas Scene (third column). Different rows represent results of using different training sizes. Specifically, the first to fourth row are the performance when the training size is 5, 10, 20 and 30 samples per class, respectively.}
\label{fig:parameters}
\end{figure}

\begin{figure}[!htbp]
\centering
\centerline{\includegraphics[width=8.50cm]{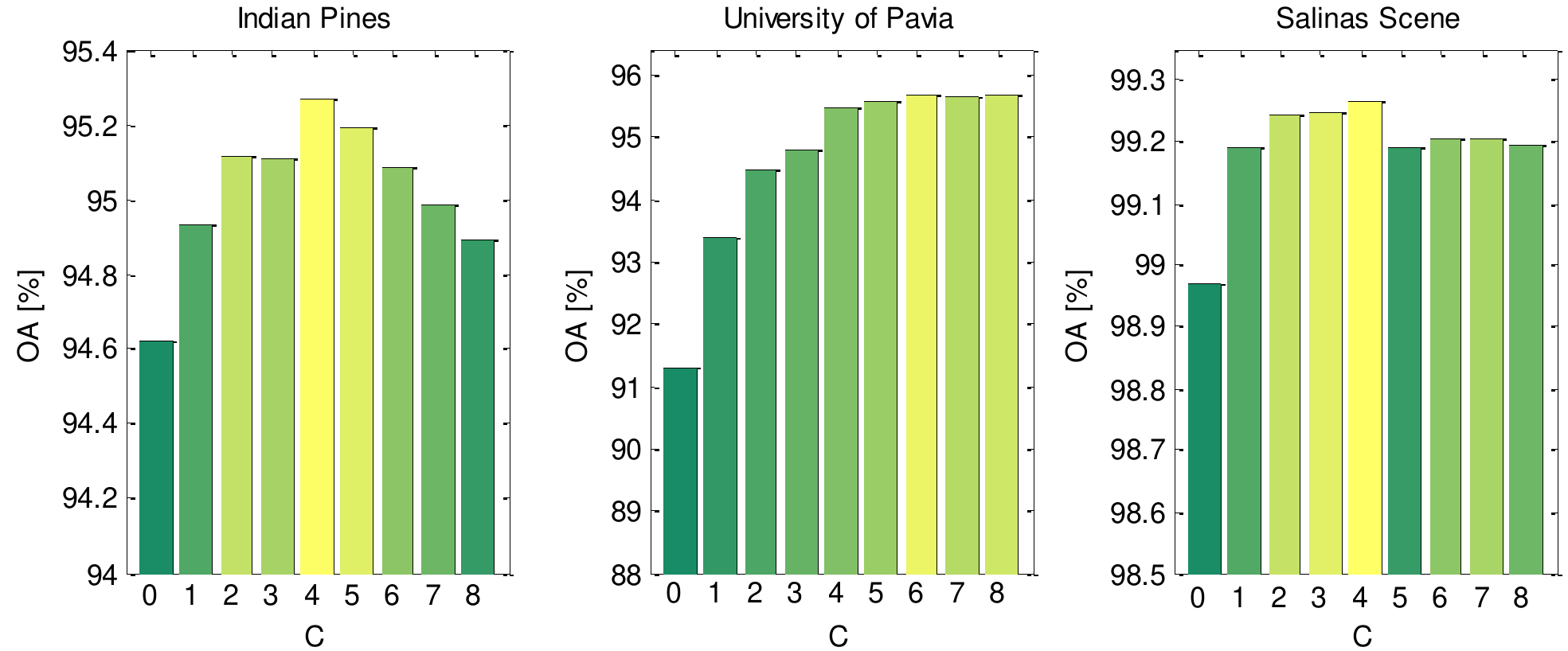}}
\caption{The OA results of the proposed approaches according to the scale number of $C$ for Indian Pines, University of Pavia, and Salinas Scene. The best performance is achieved when $C$ is set to 4, 6, and 4, for these three datasets respectively.}
\label{fig:scales}
\end{figure}

\subsection{Parameter Tuning}
In this subsection, we investigate the influences of (i) the number of superpixels in the SuperPCA approach, (ii) the number of \textcolor[rgb]{0.00,0.00,0.00}{scales} of the proposed Multiscale SuperPCA, \emph{i.e.}, the value of the power exponent in Eq. (\ref{eq:multiseg}) on the performance of the proposed SuperPCA method. Fig. \ref{fig:parameters} illustrates the OA of SuperPCA as a function of the number of superpixels, $S_f$, whose value is chosen from \{1, 3, 5, 10, 20, 30, 40, 50, 75, 100, 150, 200, 300\}. From the parameter tuning results, we can at least draw the following two conclusions:
\begin{itemize}
  \item With the increase of the number of superpixels, it shows that the overall performance will first ascend and then descend. Too large or too small number of superpixel will lead to reduced performance of the proposed SuperPCA method. This is mainly because that too large number of superpixel will result in over-segmented regions and cannot make full use of all the samples belong to the homogeneous area, while a too small number of superpixel will result in under-segmentation and introduce some samples from different homogeneous areas. \textcolor[rgb]{0.00,0.00,0.00}{In addition, when the number of superpixel is too large, each region will have a limited number of pixels, it will not guarantee the stability and reliability of the PCA results, i.e., limited number of samples are not enough to ensure that the real projection.}
  \item By setting a proper value of the number of superpixels, the performance is always better than when the number of superpixels is set to 1 (which reduces to the case of traditional global PCA method). It is evident that the proposed SuperPCA, which takes the spatial homogeneity of HSIs into account, is much more effective than the traditional PCA for capturing the intrinsic data structure.

\end{itemize}

\setlength{\tabcolsep}{2.75pt}
\begin{table*}[p]
\scriptsize
      \caption{Performance of the proposed SuperPCA approach on the Indian Pines dataset with different segmentation scales (-5 to 5).}
      \centering
        \begin{tabular}[ht]{ |c|| c | c | c  | c| c | c | c  | c| c | c | c|  }
\hline
Class Names	&	$c$ = -5	&	$c$ = -4	&	$c$ = -3	&	$c$ = -2	&	$c$ = -1	&	$c$ = 0	&	$c$ = 1	&	$c$ = 2	&	$c$ = 3	&	$c$ = 4	&	$c$ = 5	\\
\hline
\hline
 \multicolumn{1}{>{\columncolor{indian1}}c}{Alfalfa}	&	97.83 	&	96.96 	&	96.52 	&	96.96 	&	\textbf{100} 	&	\textbf{100} 	&	\textbf{100} 	&	99.13 	&	99.13 	&	99.13 	&	99.13 	\\
 \multicolumn{1}{>{\columncolor{indian2}}c}{Corn-notill}	&	83.13 	&	84.48 	&	87.83 	&	87.03 	&	\textbf{91.12} 	&	89.67 	&	88.04 	&	82.85 	&	77.52 	&	72.38 	&	60.67 	\\
 \multicolumn{1}{>{\columncolor{indian3}}c}{Corn-mintill}	&	77.95 	&	82.23 	&	87.80 	&	86.23 	&	86.18 	&	\textbf{92.44} 	&	88.88 	&	87.63 	&	85.29 	&	81.90 	&	78.30 	\\
 \multicolumn{1}{>{\columncolor{indian4}}c}{Corn}	&	91.64 	&	94.93 	&	92.56 	&	93.43 	&	94.01 	&	95.51 	&	96.28 	&	\textbf{97.63} 	&	96.23 	&	91.84 	&	88.41 	\\
 \multicolumn{1}{>{\columncolor{indian5}}c}{Grass-pasture}	&	94.92 	&	96.36 	&	95.92 	&	96.42 	&	\textbf{97.00} 	&	96.56 	&	95.74 	&	95.65 	&	94.02 	&	90.77 	&	89.89 	\\
 \multicolumn{1}{>{\columncolor{indian6}}c}{Grass-trees}	&	98.47 	&	\textbf{99.61} 	&	98.49 	&	98.57 	&	98.47 	&	97.93 	&	96.99 	&	94.93 	&	92.51 	&	82.19 	&	78.50 	\\
 \multicolumn{1}{>{\columncolor{indian7}}c}{Grass-pasture-mowed}	&	95.71 	&	94.29 	&	89.29 	&	90.00 	&	\textbf{97.14} 	&	97.14 	&	97.14 	&	97.14 	&	97.14 	&	97.14 	&	97.14 	\\
 \multicolumn{1}{>{\columncolor{indian8}}c}{Hay-windrowed}	&	99.98 	&	\textbf{100} 	&	99.80 	&	99.80 	&	\textbf{100} 	&	99.64 	&	99.64 	&	99.64 	&	99.64 	&	96.38 	&	93.62 	\\
 \multicolumn{1}{>{\columncolor{indian9}}c}{Oats}	&	94.00 	&	99.00 	&	99.00 	&	98.00 	&	\textbf{100} 	&	\textbf{100} 	&	\textbf{100} 	&	\textbf{100} 	&	\textbf{100} 	&	\textbf{100} 	&	\textbf{100} 	\\
 \multicolumn{1}{>{\columncolor{indian10}}c}{Soybean-notill}	&	84.71 	&	85.12 	&	91.09 	&	\textbf{91.40} 	&	91.19 	&	90.69 	&	90.67 	&	90.10 	&	85.22 	&	77.14 	&	69.72 	\\
 \multicolumn{1}{>{\columncolor{indian11}}c}{Soybean-mintill} 	&	90.76 	&	93.30 	&	91.18 	&	93.97 	&	94.72 	&	94.48 	&	94.08 	&	96.02 	&	\textbf{96.51} 	&	89.61 	&	88.34 	\\
 \multicolumn{1}{>{\columncolor{indian12}}c}{Soybean-clean}	&	84.74 	&	90.50 	&	91.08 	&	89.40 	&	90.48 	&	92.97 	&	92.13 	&	\textbf{94.65} 	&	93.36 	&	89.17 	&	79.98 	\\
 \multicolumn{1}{>{\columncolor{indian13}}c}{Wheat}	&	99.20 	&	99.31 	&	\textbf{99.43} 	&	\textbf{99.43} 	&	\textbf{99.43} 	&	\textbf{99.43} 	&	\textbf{99.43} 	&	\textbf{99.43} 	&	98.46 	&	98.29 	&	96.46 	\\
 \multicolumn{1}{>{\columncolor{indian14}}c}{Woods}	&	98.85 	&	98.85 	&	98.84 	&	\textbf{98.94} 	&	98.76 	&	98.89 	&	95.33 	&	91.21 	&	83.65 	&	83.31 	&	72.69 	\\
 \multicolumn{1}{>{\columncolor{indian15}}c}{Buildings-Grass-Trees-Drives}	&	97.61 	&	98.06 	&	98.43 	&	98.62 	&	97.33 	&	\textbf{98.65} 	&	98.60 	&	98.26 	&	97.36 	&	94.58 	&	90.73 	\\
 \multicolumn{1}{>{\columncolor{indian16}}c}{Stone-Steel-Towers} 	&	97.14 	&	97.30 	&	96.98 	&	97.14 	&	98.41 	&	98.41 	&	98.89 	&	99.52 	&	\textbf{99.52} 	&	98.57 	&	98.57 	\\
\hline
\hline
OA[\%]	&	90.37 	&	92.15 	&	93.01 	&	93.45 	&	94.26 	&	\textbf{94.62} 	&	93.41 	&	92.49 	&	89.84 	&	84.83 	&	79.30 	\\
AA[\%]	&	92.92 	&	94.39 	&	94.64 	&	94.71 	&	95.89 	&	\textbf{96.40} 	&	95.74 	&	95.24 	&	93.47 	&	90.15 	&	86.38 	\\
Kappa	&	0.8899 	&	0.9101 	&	0.9200 	&	0.9251 	&	0.9342 	&	\textbf{0.9383} 	&	0.9243 	&	0.9133 	&	0.8821 	&	0.8241 	&	0.7579 	\\
\hline
\end{tabular}
\label{table:Indian_scale}
\end{table*}

\setlength{\tabcolsep}{2.75pt}
\begin{table*}[p]
\scriptsize
      \caption{Performance of the proposed approach on the University of Pavia dataset with different segmentation scales (-5 to 5).}
      \centering
        \begin{tabular}[ht]{ |c|| c | c | c  | c| c | c | c  | c| c | c | c|  }
\hline
Class Names	&	$c$ = -5	&	$c$ = -4	&	$c$ = -3	&	$c$ = -2	&	$c$ = -1	&	$c$ = 0	&	$c$ = 1	&	$c$ = 2	&	$c$ = 3	&	$c$ = 4	&	$c$ = 5	\\
\hline
\hline
\multicolumn{1}{>{\columncolor{paviaU1}}c}{Asphalt}	&	71.68 	&	71.99 	&	78.32 	&	79.10 	&	78.06 	&	81.40 	&	\textbf{86.80} 	&	83.46 	&	79.01 	&	82.16 	&	77.30 	\\
\multicolumn{1}{>{\columncolor{paviaU2}}c}{Bare soil}	&	74.67 	&	93.97 	&	86.32 	&	92.12 	&	91.22 	&	\textbf{94.41} 	&	92.63 	&	88.63 	&	84.32 	&	83.76 	&	77.13 	\\
\multicolumn{1}{>{\columncolor{paviaU3}}c}{Bitumen}	&	81.58 	&	89.98 	&	88.89 	&	89.96 	&	94.40 	&	\textbf{97.09} 	&	96.02 	&	95.22 	&	89.93 	&	93.33 	&	92.35 	\\
\multicolumn{1}{>{\columncolor{paviaU4}}c}{Bricks}	&	90.89 	&	92.75 	&	\textbf{92.90} 	&	86.34 	&	89.85 	&	86.21 	&	82.88 	&	78.60 	&	78.70 	&	75.80 	&	70.59 	\\
\multicolumn{1}{>{\columncolor{paviaU5}}c}{Gravel}	&	\textbf{97.89} 	&	97.59 	&	97.53 	&	97.00 	&	97.47 	&	96.65 	&	97.05 	&	96.05 	&	91.58 	&	91.59 	&	90.31 	\\
\multicolumn{1}{>{\columncolor{paviaU6}}c}{Meadows}	&	69.40 	&	89.11 	&	86.95 	&	90.86 	&	91.02 	&	\textbf{92.23} 	&	89.96 	&	90.13 	&	84.78 	&	83.20 	&	82.11 	\\
\multicolumn{1}{>{\columncolor{paviaU7}}c}{Metal sheets}	&	90.78 	&	89.50 	&	94.69 	&	90.76 	&	94.20 	&	94.55 	&	95.66 	&	95.68 	&	95.98 	&	\textbf{97.45} 	&	96.94 	\\
\multicolumn{1}{>{\columncolor{paviaU8}}c}{Shadows}	&	73.50 	&	81.07 	&	86.08 	&	88.11 	&	87.58 	&	88.16 	&	92.44 	&	\textbf{95.52} 	&	91.62 	&	91.64 	&	86.66 	\\
\multicolumn{1}{>{\columncolor{paviaU9}}c}{Trees}	&	\textbf{99.97} 	&	99.44 	&	99.64 	&	99.65 	&	97.56 	&	98.53 	&	98.85 	&	97.47 	&	97.72 	&	97.26 	&	97.04 	\\
\hline
\hline
OA[\%]	&	76.74 	&	88.69 	&	86.61 	&	89.36 	&	89.32 	&	\textbf{91.30} 	&	91.23 	&	88.83 	&	84.92 	&	84.97 	&	80.28 	\\
AA[\%]	&	83.37 	&	89.49 	&	90.15 	&	90.43 	&	91.26 	&	92.14 	&	\textbf{92.48} 	&	91.20 	&	88.18 	&	88.47 	&	85.60 	\\
Kappa	&	0.7030 	&	0.8516 	&	0.8265 	&	0.8608 	&	0.8604 	&	\textbf{0.8856} 	&	0.8851 	&	0.8547 	&	0.8057 	&	0.8064 	&	0.7485 	\\
\hline
\end{tabular}
\label{table:PaviaU_scale}
\end{table*}

\setlength{\tabcolsep}{2.75pt}
\begin{table*}[p]
\scriptsize
      \caption{Performance of the proposed approach on the Salinas Scene dataset with different segmentation scales (-5 to 5).}
      \centering
        \begin{tabular}[ht]{ |c| c | c | c  | c| c | c | c  | c| c | c | c|  }
\hline
Class Names	&	$c$ = -5	&	$c$ = -4	&	$c$ = -3	&	$c$ = -2	&	$c$ = -1	&	$c$ = 0	&	$c$ = 1	&	$c$ = 2	&	$c$ = 3	&	$c$ = 4	&	$c$ = 5	\\
\hline
\hline
 \multicolumn{1}{>{\columncolor{indian1}}c}{Brocoli\_green\_weeds\_1}	&	\textbf{100}	&	\textbf{100}	&	\textbf{100}	&	\textbf{100}	&	\textbf{100}	&	\textbf{100}	&	\textbf{100}	&	\textbf{100}	&	\textbf{100}	&	98.74	&	93.75	\\
 \multicolumn{1}{>{\columncolor{indian2}}c}{Brocoli\_green\_weeds\_2}	&	\textbf{99.99}	&	99.95	&	99.8	&	99.96	&	99.85	&	99.78	&	99.73	&	97.9	&	96.96	&	91.57	&	82.39	\\
 \multicolumn{1}{>{\columncolor{indian3}}c}{Fallow}	&	\textbf{100}	&	\textbf{100}	&	99.97	&	99.23	&	98.21	&	99.67	&	98.32	&	99.22	&	99.54	&	96.78	&	96.03	\\
 \multicolumn{1}{>{\columncolor{indian4}}c}{Fallow\_rough\_plow}	&	99.07	&	99.02	&	99.05	&	99.12	&	98.91	&	99.16	&	\textbf{99.32}	&	99.27	&	96.74	&	95.50	&	93.20	\\
 \multicolumn{1}{>{\columncolor{indian5}}c}{Fallow\_smooth}	&	99.45	&	\textbf{99.46}	&	99.44	&	99.38	&	98.69	&	99.37	&	98.62	&	98.70	&	96.72	&	95.32	&	89.13	\\
 \multicolumn{1}{>{\columncolor{indian6}}c}{Stubble}	 	&	\textbf{99.88}	&	99.87	&	99.76	&	99.82	&	99.79	&	98.37	&	98.33	&	98.68	&	94.83	&	87.45	&	80.62	\\
 \multicolumn{1}{>{\columncolor{indian7}}c}{Celery}	 	&	\textbf{99.91}	&	99.55	&	98.06	&	98.19	&	98.11	&	97.78	&	98.00	&	97.81	&	96.73	&	94.65	&	81.93	\\
 \multicolumn{1}{>{\columncolor{indian8}}c}{Grapes\_untrained}	 	&	96.63	&	93.66	&	94.87	&	96.62	&	96.52	&	99.39	&	\textbf{99.48}	&	98.13	&	98.93	&	94.74	&	97.02	\\
 \multicolumn{1}{>{\columncolor{indian9}}c}{Soil\_vinyard\_develop}	 	&	99.25	&	99.39	&	99.54	&	\textbf{99.58}	&	99.51	&	99.02	&	99.57	&	95.11	&	89.46	&	81.65	&	68.6	\\
 \multicolumn{1}{>{\columncolor{indian10}}c}{Corn\_senesced\_green\_weeds}	 	&	92.55	&	96.09	&	\textbf{97.25}	&	97.08	&	96.59	&	97.16	&	94.35	&	94.59	&	89.89	&	83.56	&	79.72	\\
 \multicolumn{1}{>{\columncolor{indian11}}c}{Lettuce\_romaine\_4wk}	 	&	98.01	&	98.30	&	98.05	&	98.14	&	98.61	&	98.38	&	98.42	&	98.72	&	\textbf{98.78}	&	95.31	&	92.36	\\
 \multicolumn{1}{>{\columncolor{indian12}}c}{Lettuce\_romaine\_5wk}	 	&	99.28	&	99.35	&	99.27	&	99.67	&	98.34	&	\textbf{99.80}	&	99.51	&	99.10	&	97.53	&	96.85	&	91.75	\\
 \multicolumn{1}{>{\columncolor{indian13}}c}{Lettuce\_romaine\_6wk}	 	&	98.21	&	98.21	&	98.21	&	98.19	&	98.23	&	\textbf{98.28}	&	98.09	&	97.99	&	97.79	&	97.81	&	97.14	\\
 \multicolumn{1}{>{\columncolor{indian14}}c}{Lettuce\_romaine\_7wk}	 	&	95.41	&	97.83	&	97.74	&	98.06	&	\textbf{98.25}	&	97.95	&	98.01	&	96.29	&	94.92	&	93.65	&	92.57	\\
 \multicolumn{1}{>{\columncolor{indian15}}c}{Vinyard\_untrained}	 	&	91.17	&	97.46	&	97.11	&	95.96	&	96.12	&	\textbf{99.05}	&	97.43	&	94.58	&	83.91	&	71.9	&	60.08	\\
 \multicolumn{1}{>{\columncolor{indian16}}c}{Vinyard\_vertical\_trellis}		&	99.18	&	\textbf{99.33}	&	98.98	&	99.18	&	98.92	&	98.99	&	98.66	&	97.97	&	91.54	&	87.30	&	85.17	\\
\hline
\hline
	OA[\%]	&	97.29	&	97.78	&	97.93	&	98.16	&	97.99	&	\textbf{98.97}	&	98.57	&	97.26	&	94.19	&	88.85	&	83.12	\\
	AA[\%]	&	98.00	&	98.59	&	98.57	&	98.64	&	98.42	&	\textbf{98.89}	&	98.49	&	97.75	&	95.27	&	91.42	&	86.34	\\
	Kappa	&	0.9698	&	0.9753	&	0.9770	&	0.9795	&	0.9776	&	\textbf{0.9886}	&	0.9841	&	0.9694	&	0.9349	&	0.8746	&	0.8088	\\
\hline
\end{tabular}
\label{table:Salinas_scale}
\end{table*}

\setlength{\tabcolsep}{2.35pt}
\begin{table*}[!htbp]
\scriptsize
      \caption{Classification results (in \textcolor[rgb]{0.00,0.00,0.00}{terms} of OA) of the proposed approaches and eight comparison algorithms on three HSI datasets with different training numbers.}
      \centering
        \begin{tabular}[ht]{ |c| c || c | c  | c| c | c | c  | c| c | c | c|  }
\hline
Datasets	&	T.N.s/C	&	Raw	&	PCA	&	ICA	&	LPP	&	NPE	&	LPNPE	&	LDA	&	LFDA	&	SuperPCA	&	MSuperPCA	\\	
\hline
\hline
��	&	5	&	44.88	&	46.37	&	45.21	&	53.58	&	53.68	&	67.25	&	59.95	&	59.62	&	77.34	&	\textbf{78.68}	\\	
Indian	&	10	&	55.77	&	55.72	&	57.12	&	70.41	&	70.49	&	76.45	&	69.30	&	64.91	&	85.76	&	\textbf{87.12}	\\	
Pines	&	20	&	63.81	&	62.97	&	64.41	&	80.26	&	79.87	&	83.51	&	76.56	&	74.01	&	93.90	&	\textbf{95.69}	\\	
��	&	30	&	68.77	&	67.27	&	68.92	&	84.43	&	83.98	&	90.10	&	89.51	&	90.19	&	94.62	&	\textbf{96.78}	\\	
��	&	200	&	84.01	&	84.40	&	82.86	&	94.31	&	94.16	&	97.80	&	98.55	&	\textbf{99.15}	&	97.13	&	98.25\\	
\hline
\hline
��	&	5	&	64.59	&	65.26	&	66.58	&	70.86	&	68.35	&	76.12	&	72.43	&	74.67	&	74.39	&	\textbf{78.49}	\\	
University	&	10	&	70.22	&	70.15	&	71.39	&	81.29	&	80.63	&	82.55	&	81.24	&	78.95	&	83.42	&	\textbf{91.67}	\\	
of Pavia	&	20	&	75.85	&	75.91	&	76.65	&	86.00	&	85.69	&	88.56	&	85.00	&	86.98	&	89.38	&	\textbf{95.37}	\\	
��	&	30	&	76.45	&	76.31	&	76.87	&	86.90	&	87.19	&	90.56	&	87.91	&	90.19	&	91.30	&	\textbf{95.68}	\\	
��	&	200	&	85.71	&	85.70	&	85.79	&	94.08	&	93.69	&	97.50	&	95.72&	98.72	&	96.99	&	\textbf{98.84}	\\
\hline
\hline
	&	5	&	81.79	&	81.87	&	81.75	&	85.23	&	84.86	&	92.09	&	89.03	&	88.83	&	94.42	&	\textbf{95.00}	\\	
Salinas	&	10	&	85.24	&	85.28	&	85.74	&	88.60	&	88.99	&	94.52	&	91.46	&	82.77	&	96.78	&	\textbf{98.15}	\\	
Scene	&	20	&	87.85	&	87.79	&	88.08	&	90.61	&	90.69	&	95.89	&	93.72	&	93.56	&	98.37	&	\textbf{99.04}	\\	
��	&	30	&	88.93	&	89.24	&	89.28	&	91.73	&	91.69	&	96.66	&	95.87	&	95.89	&	98.97	&	\textbf{99.27}	\\	
��	&	200	&	91.48	&	91.94	&	91.74	&	96.18	&	95.88	&	99.09	&	99.87	&	99.57	&	99.63	&	\textbf{99.70}	\\
\hline
\end{tabular}
\label{table:com_stateofthearts}
\end{table*}

Based on the above experiments, we can obtain the optimal fundamental superpixel number $S_f$ for Indian Pines, University of Pavia, and Salinas Scene, which is set to 100, 20, and 100, respectively.

In order to verify the necessity of multiscale fusion, we report the classification results of the proposed SuperPCA approach with different segmentation scales, \emph{i.e.}, $c$ is set from -5 to 5. When $c$ is set to zero, it means that the input HSI is segmented with the fundamental superpixel number. Tables \ref{table:Indian_scale}, \ref{table:PaviaU_scale}, and \ref{table:Salinas_scale} tabulate the OA, AA, and Kappa coefficient \textcolor[rgb]{0.00,0.00,0.00}{when the training number is 30}, under different segmentation scales for Indian Pines, University of Pavia, and Salinas Scene images, respectively. \textcolor[rgb]{0.00,0.00,0.00}{The best performance for each class is highlighted in bold typeface.} From these tables, we learn that even though SuperPCA can obtain the best overall performance when the power exponent $c$ is set to zero, \emph{i.e.}, under the fundamental superpixel number $S_f$, it cannot achieve the best performance in every class (for the sake of convenience, we highlight the best performance for each class (row) in bold). For example, as shown in Table \ref{table:Indian_scale}, when $c=-4$, the OA is obviously inferior to the best scale, \emph{i.e.} 92.15\% to 94.62\%. In this case, however, the sixth and eighth classes get the best classification accuracy. Similar results can be also observed in Table \ref{table:PaviaU_scale} and Table \ref{table:Salinas_scale} (please refer to the case when $c=-5$). The OA is minimum, but it achieves the best classification performances in the fifth and ninth classes. All these results demonstrate that the superpixel segmentation based on a single scale is not able to fully model the complexity and diversity of HSIs. Therefore, it is an effective and reliable choice to perform multiscale segmentation based decision fusion for HSI classification.

To further verify the usefulness of the multiscale segmentation strategy as well as to assess the influence of the different values of $C$, as shown in Fig.~\ref{fig:scales}, we show the OA results of the proposed multiscale SuperPCA method according to scale number $C$ for the images of Indian Pines, University of Pavia, and Salinas Scene. By fusing multiscale segmentation based classification results, we can expect better results than single scale SuperPCA method, \emph{i.e.}, setting $C$ to 0. When setting the value of $C$ to 4, 6, and 4 for Indian Pines, University of Pavia, and Salinas Scene images, the improvements of single scale SuperPCA method over multiscale SuperPCA method are 0.65\%, 4.38\%, and 0.30\%, respectively. From these results, we observe that the improvement on the University of Pavia image is more obvious than the other two images. This is mainly due to the following two reasons: on the one hand, the single scale SuperPCA performs not very well and has a relatively large space for improvement. On the other hand, the University of Pavia image has richer and more complex texture information, and it is much more difficult for the single scale based segmentation method to capture these useful spatial knowledge. At the same time, we also observe another phenomenon: in order to achieve high classification accuracy, the relatively complex HSIs may require a larger scale number to exploit its spatial information, \emph{e.g.}, the optimal scale number of the University of Pavia image is 6, which is larger than that of the other two HSI datasets.

\begin{figure*}[t]
\centering
\centerline{\includegraphics[width=15.20cm]{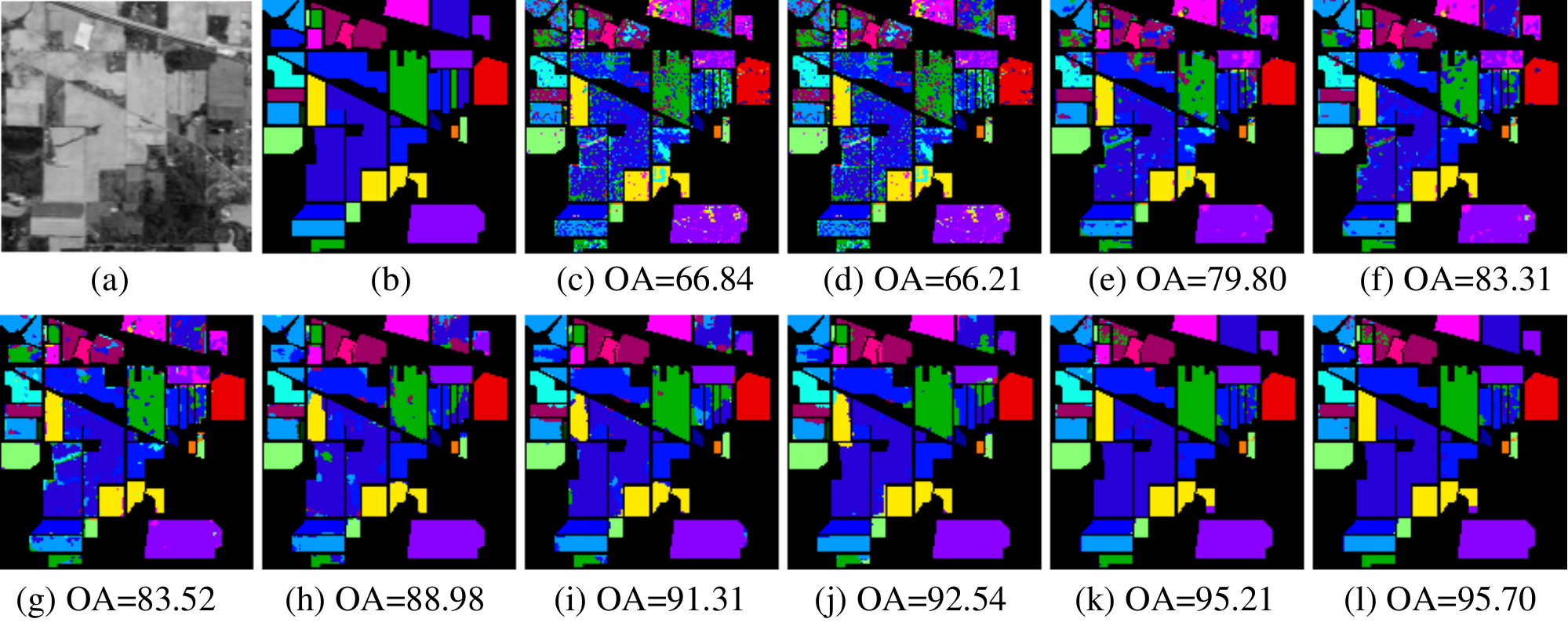}}
\vspace{-0.150cm}
\caption{Classification maps obtained with the Indian Pines dataset. (a) First principal component, (b) Ground truth, (c) Raw pixel, (d) PCA~\cite{scholkopf1998nonlinear}, (e) ICA \cite{wang2006independent}, (f) LPP~\cite{he2004locality}, (g) NPE~\cite{he2005neighborhood}, (h) LPNPE~\cite{zhou2015dimension}, (i) LDA~\cite{prasad2008limitations}, (f) LFDA ~\cite{li2012locality}, (k) SuperPCA, (l) MSuperPCA.}
\label{fig:map_Indian}
\end{figure*}

\begin{figure*}[t]
\centering
\centerline{\includegraphics[width=15.20cm]{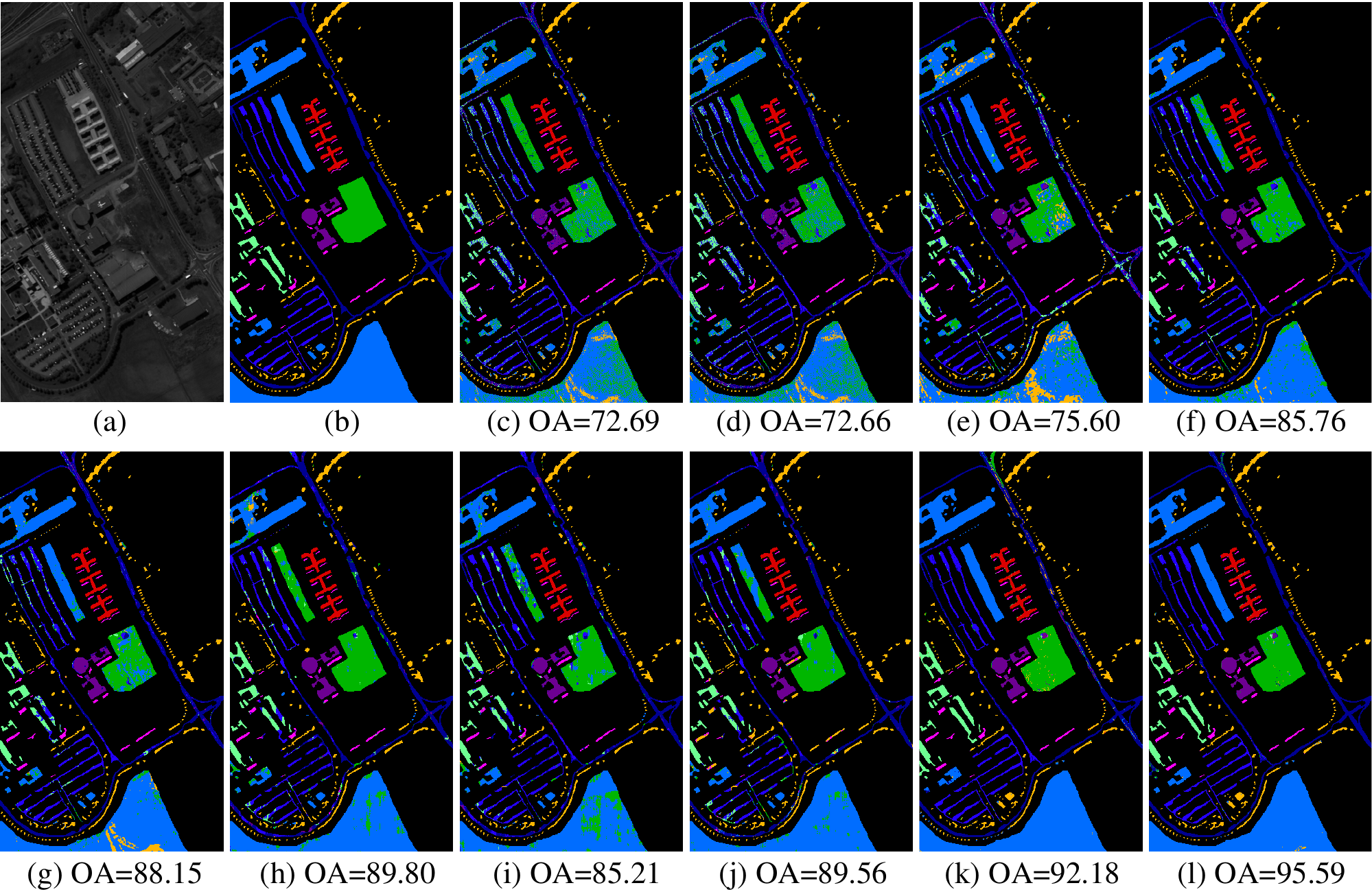}}
\vspace{-0.150cm}
\caption{Classification maps obtained with the University of Pavia dataset. (a) First principal component, (b) Ground truth, (c) Raw pixel, (d) PCA~\cite{scholkopf1998nonlinear}, (e) ICA \cite{wang2006independent}, (f) LPP~\cite{he2004locality}, (g) NPE~\cite{he2005neighborhood}, (h) LPNPE~\cite{zhou2015dimension}, (i) LDA~\cite{prasad2008limitations}, (f) LFDA ~\cite{li2012locality}, (k) SuperPCA, (l) MSuperPCA.}
\label{fig:map_PaviaU}
\end{figure*}

\begin{figure*}[t]
\centering
\centerline{\includegraphics[width=15.20cm]{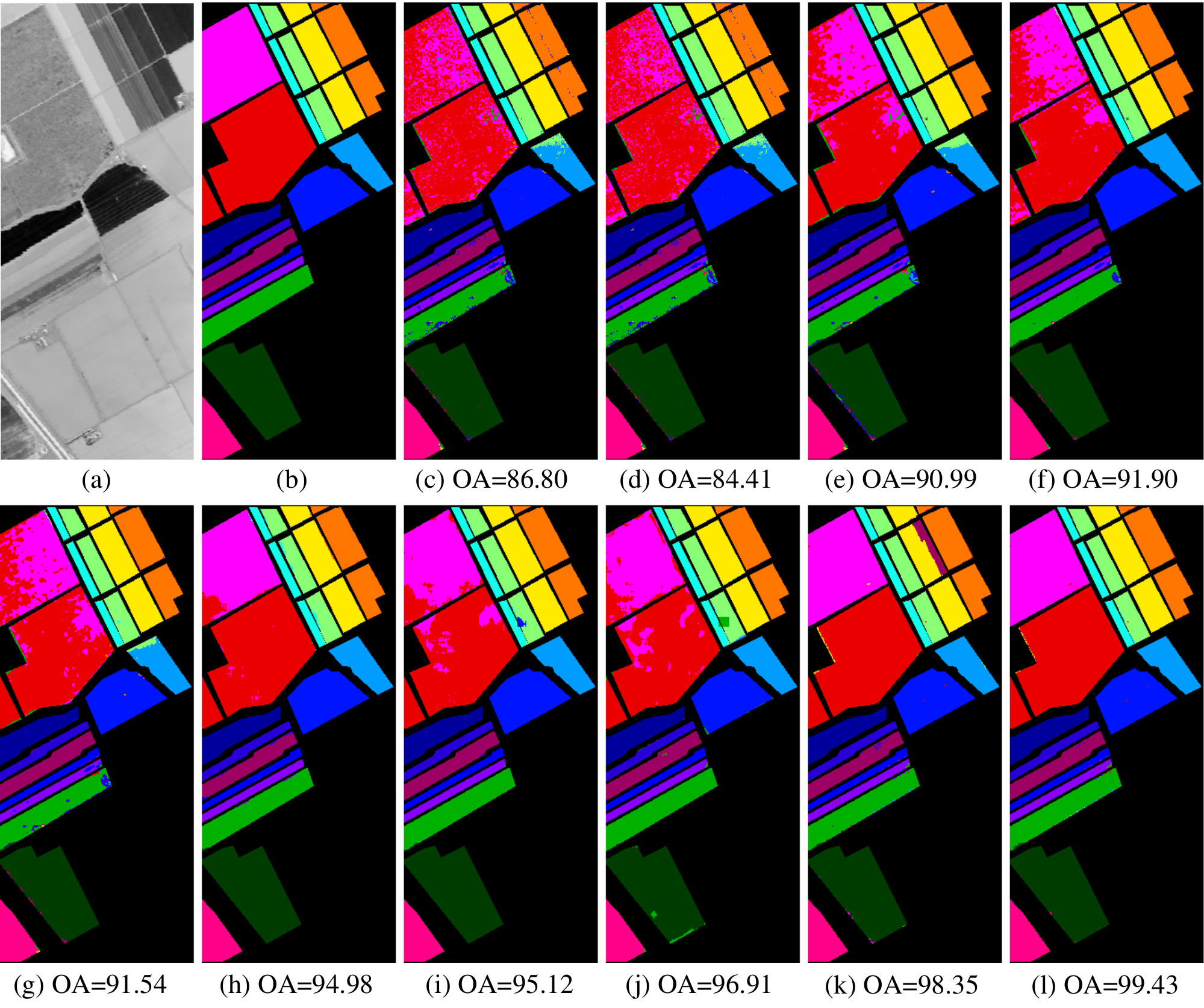}}
\vspace{-0.150cm}
\caption{Classification maps obtained with the Salinas Scene dataset. (a) First principal component, (b) Ground truth, (c) Raw pixel, (d) PCA~\cite{scholkopf1998nonlinear}, (e) ICA \cite{wang2006independent}, (f) LPP~\cite{he2004locality}, (g) NPE~\cite{he2005neighborhood}, (h) LPNPE~\cite{zhou2015dimension}, (i) LDA~\cite{prasad2008limitations}, (f) LFDA ~\cite{li2012locality}, (k) SuperPCA, (l) MSuperPCA.}
\label{fig:map_Salinas}
\end{figure*}

\subsection{Comparison Results with State-of-the-arts}
\textcolor[rgb]{0.00,0.00,0.00}{The classification maps obtained with above-mentioned three public HSI datasets for the proposed SuperPCA and MSuperPCA approaches and the comparison methods are given in Fig. \ref{fig:map_Indian}, Fig. \ref{fig:map_PaviaU}, and Fig. \ref{fig:map_Salinas}. Here, we only show the results when the number of training samples is set to 30. From these maps, we can learn that raw spectral features based method, PCA~\cite{scholkopf1998nonlinear}, ICA \cite{wang2006independent}, LPP~\cite{he2004locality}, and NPE~\cite{he2005neighborhood} exhibit higher classification errors than other methods. Among the comparison unsupervised methods, LPNPE~\cite{zhou2015dimension} achieves the best performance due to its  local spatial¨Cspectral scatter based effective spatial information extraction. By taking advantage of the discrimination information of the labeled samples, these supervised methods (LDA~\cite{prasad2008limitations} and LFDA ~\cite{li2012locality}) can produce very good results. As for the datasets of Indian Pines and University of Pavia, the proposed SuperPCA and MSuperPCA are clearly better than previous arts. When comparing the classification maps of our methods with LDA~\cite{prasad2008limitations} and LFDA ~\cite{li2012locality} (please refer to the University of Pavia dataset), it can be observed that our methods can achieve much better results for these large regions (i.e., Bare soil and Meadows), which can be attributed to the efficient segmentation. Through the fusion of multiscale segmentation based classification results, MSuperPCA can improve the result of single scale segmentation based SuperPCA and obtain accurate classification maps (please refer to the edges and the holes of regions).}

\textcolor[rgb]{0.00,0.00,0.00}{According to the experimental settings of LPNPE method~\cite{zhou2015dimension}, which can be seen as the best unsupervised feature extraction method for HSI classification to the best of our knowledge, we further randomly choose $T=5,10,20,30$ samples from each class to form the training set to test the comparison results (in terms of OA), respectively. Table~\ref{table:com_stateofthearts} tabulates the OA performance of different approaches.}
From the results of each individual method, the OA performance of the proposed SuperPCA is better than others in most instances, and this advantage is particularly evident when the number of training samples is small. With the increase of training number, the performance of supervised methods becomes much better. This can be explained as follows: with the increase of labeled training samples, these supervised methods are able to use more discriminant information from the training samples.

\textcolor[rgb]{0.00,0.00,0.00}{To further utilize the spatial information of HSIs, MSuperPCA is advocated to fuse the decisions of SuperPCA with different segmentation scales. By comparing the last two columns, we can clearly see that the performance of MSuperPCA is better than that of SuperPCA in all cases (regardless of different datasets or different training sample numbers). In particular, the improvement of MSuperPCA over SuperPCA is much more impressive on the University of Pavia dataset, that is over 4\% higher accuracy. It is because of the rich texture information contained in that dataset}.

\textcolor[rgb]{0.00,0.00,0.00}{The above results show that our proposed method can achieve good performance when the number of training samples are small. At the fourth row of each block}, we additionally provide the results when \textcolor[rgb]{0.00,0.00,0.00}{the number of training samples is relatively} large, \emph{e.g.}, $T=200$. 
In this situation, these supervised methods (LDA~\cite{prasad2008limitations} and LFDA ~\cite{li2012locality}) can learn more discriminative information from the labeled training data for classification. Therefore, all of them have considerable performance improvements as compared to the situation when the training samples are limited. \textcolor[rgb]{0.00,0.00,0.00}{Nevertheless, the results of SuperPCA, which do not use any label information, are still very competitive in this situation. By fusing multiscale classification results, MSuperPCA can even surpass LDA~\cite{prasad2008limitations} and LFDA ~\cite{li2012locality} on the University of Pavia and Salinas datasets. This proves the effectiveness of the proposed method once again.}

\subsection{Running Times}
In Table \ref{table:runningtimes}, we report the run times of extracting the dimensionality reduced features of different algorithms on the Indian Pines, University of Pavia, and Salinas Scene images with different training numbers ($T=5, 10, 20, 30$). \textcolor[rgb]{0.00,0.00,0.00}{As for the proposed method, we report the whole running time including the segmentation and \textcolor[rgb]{0.00,0.00,0.00}{dimensionality} reduction of all superpixels.} All methods were tested on MATLAB R2014a using an Intel Xeon CPU with 3.50 GHz and 16G memory PC with Windows platform. The testing time of all methods is measured using a single-threaded MATLAB process. It should be noted that for these unsupervised methods, PCA~\cite{scholkopf1998nonlinear}, ICA \cite{wang2006independent}, \textcolor[rgb]{0.00,0.00,0.00}{LPP~\cite{he2004locality}, NPE~\cite{he2005neighborhood}, LPNPE~\cite{zhou2015dimension}}, and our proposed SuperPCA, the running time will not change with the training number per class. PCA, LDA~\cite{prasad2008limitations}, and LFDA ~\cite{li2012locality} show the fastest performance. \textcolor[rgb]{0.00,0.00,0.00}{While LPP~\cite{he2004locality} and NPE~\cite{he2005neighborhood} need to construct the large similarity graph and decompose it via SVD, and thus the computational complexities of them are relatively high.}  With the increase of training numbers, the run times of these supervised methods (LDA~\cite{prasad2008limitations} and LFDA ~\cite{li2012locality}) will also increase. The timings reveal that although our method requires pre-segmentation and feature extraction for each region, the run time is still acceptable. Also, thanks to the independence of the dimensionality reduction of each superpixel, we can accelerate the algorithm simply by parallel computation.

\setlength{\tabcolsep}{1.4pt}
\begin{table}[!htbp]
\scriptsize
\caption{Running times of the feature extraction process (in seconds) of the proposed approach and some comparison algorithms on the three HSI datasets with different training numbers.}
      \centering
        \begin{tabular}[ht]{ |c| c | c | c  | c| c | c | c  | c  | c |  }
\hline
Dataset	&	T.N.s/C	&	PCA	&	ICA	&	LPP	&	NPE	&	LPNPE	&	LDA	&	LFDA	&	SuperPCA	\\
\hline
\hline
	&	5	&	0.0076 	&	2.5417 	&  0.2428 	&0.7018 	&0.3834 	&	0.0027 	&	0.0167 	&	0.6879 	\\
Indian	&	10	&	0.0076 	&	2.5417 	&0.2428 	&0.7018 	&0.3834 	&	0.0047 	&	0.0249 	&	0.6879 	\\
Pines	&	20	&	0.0076 	&	2.5417 	&0.2428 	&0.7018 	&0.3834 	&	0.0054 	&	0.0388 	&	0.6879 	\\
	&	30	&	0.0076 	&	2.5417 	&0.2428 	&0.2428 	&0.3834 	&	0.0057 	&	0.0356 	&	0.6879 	\\
\hline
	&	5	&	0.4004   &   3.1445 	& 2.9477 	&6.5322 	&	1.2357 	 	&	0.0017 	&	0.0053 	&	2.8867 	\\
University	&	10	&	0.4004 	& 3.1445 	& 2.9477  	&6.5322 	&	1.2357 	&	0.0022 	&	0.0110 	&	2.8867 	\\
of Pavia	&	20	&	0.4004 	&   3.1445 	& 2.9477  	&6.5322 	&	1.2357 	&	0.0024 	&	0.0097 	&	2.8867 	\\
	&	30	&	0.4004  &3.1445 	& 2.9477    	&6.5322 	&	1.2357 		&	0.0024 	&	0.0095 	&	2.8867 	\\
\hline
	&	5	&	0.4145   & 5.8702 	& 4.7492 	&    9.8365 	&	1.2003 	&	0.0027 	&	0.0174 	&	2.7452 	\\
Salinas	&	10	&	0.4145   & 5.8702 	& 4.7492 	&    9.8365 	&	1.2003 	&	0.0046 	&	0.0260 	&	2.7452 	\\
Scene	&	20	&	0.4145 & 5.8702 	& 4.7492 	&    9.8365 	&	1.2003 	&	0.0066 	&	0.0377 	&	2.7452 	\\
	&	30	&	0.4145  & 5.8702 	& 4.7492 	&    9.8365 	&	1.2003 	&	0.0067 	&	0.0391 	&	2.7452 	\\
\hline
\end{tabular}
\label{table:runningtimes}
\end{table}

\subsection{\textcolor[rgb]{0.00,0.00,0.00}{Discussions}}
\textcolor[rgb]{0.00,0.00,0.00}{Since the key idea of the proposed methods is to oversegement the HSIs and perform PCA superpixelwisely, how to determine the parameter of the superpixel segmentation model (i.e., the superpixel number in ERS) and the segmentation scales is a a crucial and open problem. In this paper, we set them experimentally to achieve the best performance. In fact, the segmentation scales and superpixel number jointly determine the minimum and maximum homogeneous regions, which can be deduced from the Eq. (\ref{eq:multiseg}). The searching of optimal segmentation scales and superpixel number can be converted to the problem of setting the size of minimum and maximum homogeneous regions of the given HSIs. Obviously, the size of homogeneous region is determined by the texture information. Therefore, the most direct approach is to detect the edges in a given images through some edge detectors, such as Canny and Sobel. Therefore, we can obtain the texture ratio, which can be used to define the size of homogeneous region.}

\section{Conclusions}
\label{sec:conclusions}
In this paper, we propose a \textcolor[rgb]{0.00,0.00,0.00}{simple but very effective} technique for unsupervised feature extraction of hyperspectral imagery based on superpixelwise principal component analysis (SuperPCA).
By segmenting the entire hyperspectral image (HSI) to many different homogeneous regions, which have similar reflectance \textcolor[rgb]{0.00,0.00,0.00}{properties}, it can facilitate the dimensionality reduction process of finding the essential low-dimensional feature space of HSIs. To take full advantage of the spatial information \textcolor[rgb]{0.00,0.00,0.00}{contained in the HSIs, which cannot be extracted using a single scale,} we further advocate a decision fusion strategy through multiscale segmentation based on the SuperPCA model (MSuperPCA).
Extensive experiments on three standard HSI datasets demonstrate that the proposed SuperPCA and MSuperPCA algorithms outperform the existing state-of-the-art feature extraction methods, including unsupervised feature extraction methods as well as supervised feature extraction methods, especially when the training samples are limited. \textcolor[rgb]{0.00,0.00,0.00}{When the number of the training samples is  relatively large, the proposed algorithm can still obtain very competitive classification results when compared with these supervised feature extraction methods.} Because of inheriting the merits of PCA technology, the proposed SuperPCA can be also used as a preprocessing for many hyperspectral image processing and analysis tasks.


\ifCLASSOPTIONcaptionsoff
  \newpage
\fi

{
\bibliographystyle{IEEEtran}
\bibliography{JStars2017}
}
\end{document}